\pdfoutput=1
\documentclass[11pt, letterpaper, logo, onecolumn, copyright, numbering]{paper_improved}

\usepackage{soul}
\usepackage{xcolor}

\newcommand{\hlgreen}[1]{\sethlcolor{green!30}\hl{#1}}
\newcommand{\hlblue}[1]{\sethlcolor{blue!20}\hl{#1}}
\newcommand{\hlred}[1]{\sethlcolor{red!20}\hl{#1}}

\newcommand{\hlyellow}[1]{\sethlcolor{yellow!50}\hl{#1}}
\usepackage{booktabs}
\usepackage{enumitem}
\usepackage{array}
\usepackage{colortbl}
\usepackage{xcolor}
\usepackage{caption}
\usepackage{tabularray}
\usepackage[utf8]{inputenc} 
\usepackage[T1]{fontenc}    
\usepackage{hyperref}       
\usepackage{url}            
\usepackage{booktabs}       
\usepackage{amsfonts}       
\usepackage{nicefrac}       
\usepackage{microtype}      
\usepackage{xcolor}         
\usepackage{graphicx}

\usepackage{natbib}
\usepackage{verbatim}
\usepackage{inconsolata}
\usepackage{tabularx}
\usepackage{float}
\usepackage{algorithmic}
\usepackage{multirow}
\usepackage{listings}
\usepackage{xcolor}
\usepackage{subfigure}
\usepackage{subcaption}
\usepackage{amsmath}
\usepackage{amssymb}
\usepackage{array}
\usepackage[ruled,linesnumbered]{algorithm2e}
\usepackage[T1]{fontenc}
\definecolor{lightgray}{rgb}{0.95, 0.95, 0.95}
\definecolor{darkgray}{rgb}{0.4, 0.4, 0.4}
\definecolor{backcolour}{rgb}{0.95,0.95,0.92}
\definecolor{myblue}{rgb}{0.2, 0.4, 0.8} 
\definecolor{mygreen}{rgb}{0.2, 0.6, 0.2} 

\usepackage{amsmath,amssymb,amsfonts}

\let\cite\citep

\title{Marco-LLM: Bridging Languages via Massive Multilingual Training for Cross-Lingual Enhancement}

\reportnumber{} 

\author[*,1]{Lingfeng Ming*, Bo Zeng*, Chenyang Lyu*, Tianqi Shi, Yu Zhao, Xue Yang, Yefeng Liu, Yiyu Wang, Linlong Xu, Yangyang Liu, Xiaohu Zhao, Hao Wang, Heng Liu, Hao Zhou, Huifeng Yin, Zifu Shang, Haijun Li, Longyue Wang$^\dagger$, Weihua Luo, Kaifu Zhang\\ ~ \\ 
\bf MarcoPolo Team, Alibaba International Digital Commerce}

\footnotetext[1]{Equal Contribution.}
\footnotetext[2]{Corresponding Author: wanglongyue.wly@alibaba-inc.com}

\begin{abstract}
Large Language Models~(LLMs) have achieved remarkable progress in recent years; however, their excellent performance is still largely limited to major world languages, primarily English. Many LLMs continue to face challenges with multilingual tasks, especially when it comes to low-resource languages. 
To address this issue, we introduced Marco-LLM: \hlblue{Ma}ssive multilingual t\hlgreen{r}aining for \hlred{c}r\hlyellow{o}ss-lingual enhancement LLM.
We have collected a substantial amount of multilingual data for several low-resource languages and conducted extensive continual pre-training using the Qwen2 models. This effort has resulted in a multilingual LLM named Marco-LLM. Through comprehensive evaluations on various multilingual benchmarks, including MMMLU, AGIEval, Belebele, Flores-200, XCOPA and many others, Marco-LLM has demonstrated substantial improvements over state-of-the-art LLMs. Furthermore, Marco-LLM achieved substantial enhancements in any-to-any machine translation tasks, showing the effectiveness of our multilingual LLM. 
Marco-LLM is a pioneering multilingual LLM designed to not only perform exceptionally well in multilingual tasks, including low-resource languages, but also maintain strong performance in English and other major languages, closing the performance gap between high- and low-resource language capabilities. By bridging languages, this effort demonstrates our dedication to ensuring LLMs work accurately across various languages.

\end{abstract}

\begin{document}

\maketitle

\begin{figure}[h]
    \centering
    \includegraphics[width=0.87\linewidth]{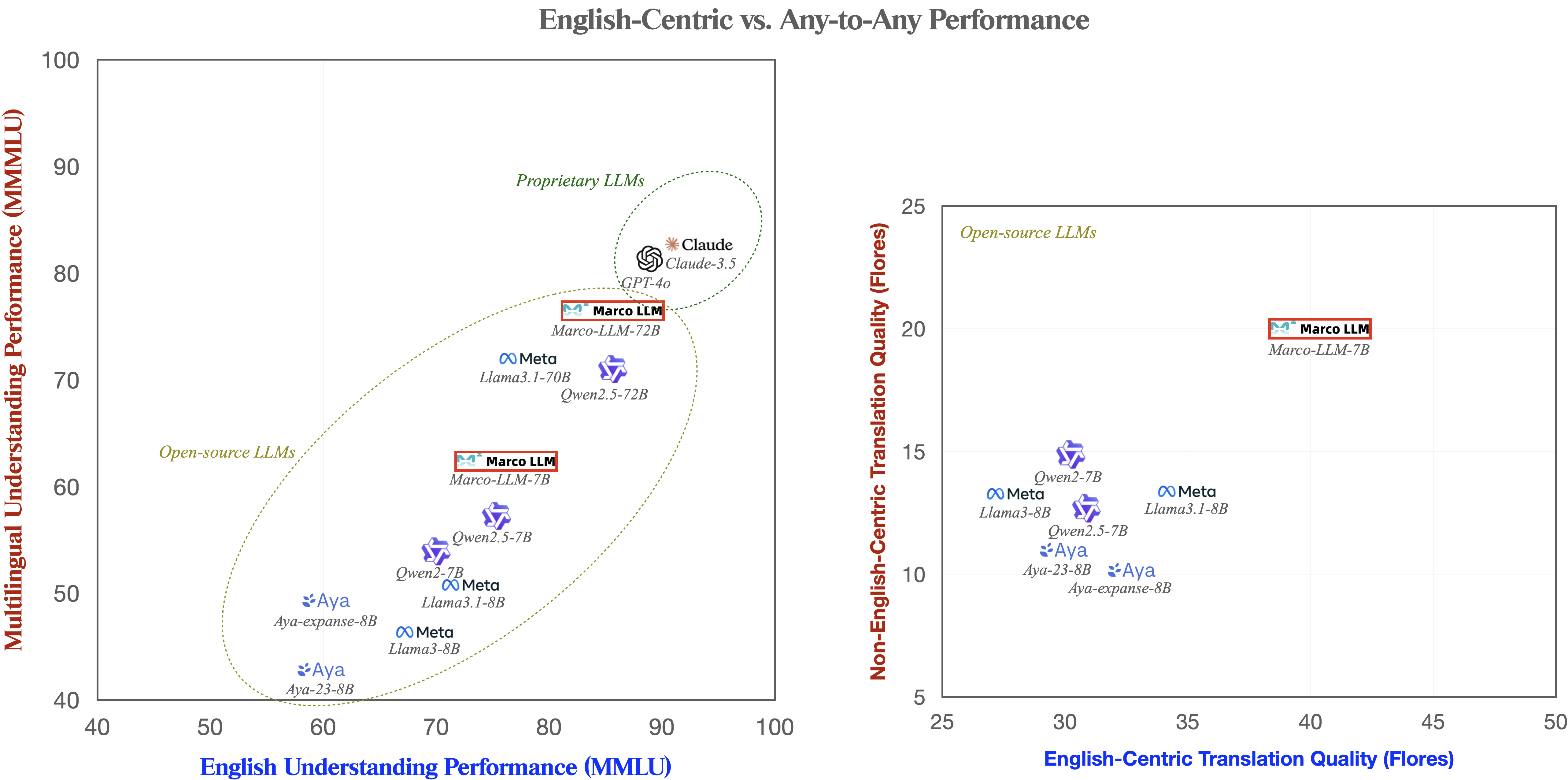}
    \caption{Comparison of English-centric performance vs Multilingual performance on MMMLU and Flores. Our Marco-LLM demonstrates strong performance on both dimensions.}
    \label{fig:vs}
\end{figure}

\cleardoublepage
\phantomsection
\tableofcontents

\newpage
\section{Introduction}

Large Language Models~(LLMs)~\cite{brown2020language_gpt3,openai2023gpt4,touvron2023llamaopenefficientfoundation,touvron2023llama2openfoundation,dubey2024llama3herdmodels,guo2024deepseekcoderlargelanguagemodel} have transformed the field of Natural Language Processing (NLP) by achieving impressive results across a variety of tasks such as language understanding, generation, and translation~\cite{bang2023multitask_chatgpt_eval,wang2023document,jiao2023chatgpt,bai2023qwen,hui2024qwen25codertechnicalreport,yao2024tree_tot}. Models like GPT-4~\cite{openai2023gpt4}, GPT-4o~\cite{hurst2024gpt4o}, PaLM\cite{Chowdhery2023palm}, and LLaMA~\cite{dubey2024llama3herdmodels} have demonstrated that scaling up model size and training data leads to significant performance gains. However, the majority of these advances have been centered around high-resource languages, predominantly English~\cite{bang2023multitask_chatgpt_eval,jiao2023chatgpt,lai2023chatgpt_English}. This focus has resulted in a performance gap when these models are applied to multilingual tasks, especially involving low-resource languages.

Multilingual NLP faces unique challenges due to the diversity and imbalance of linguistic resources~\cite{pires2019multilingual_mbert,conneau2019cross_xlm}. Low-resource languages often lack the extensive textual data required for training large models, which hinders the development of effective language technologies for these languages. As a result, speakers of low-resource languages are underrepresented in the benefits brought by recent advancements in NLP.

\begin{figure}[t]
    \centering
    \includegraphics[width=0.83\linewidth]{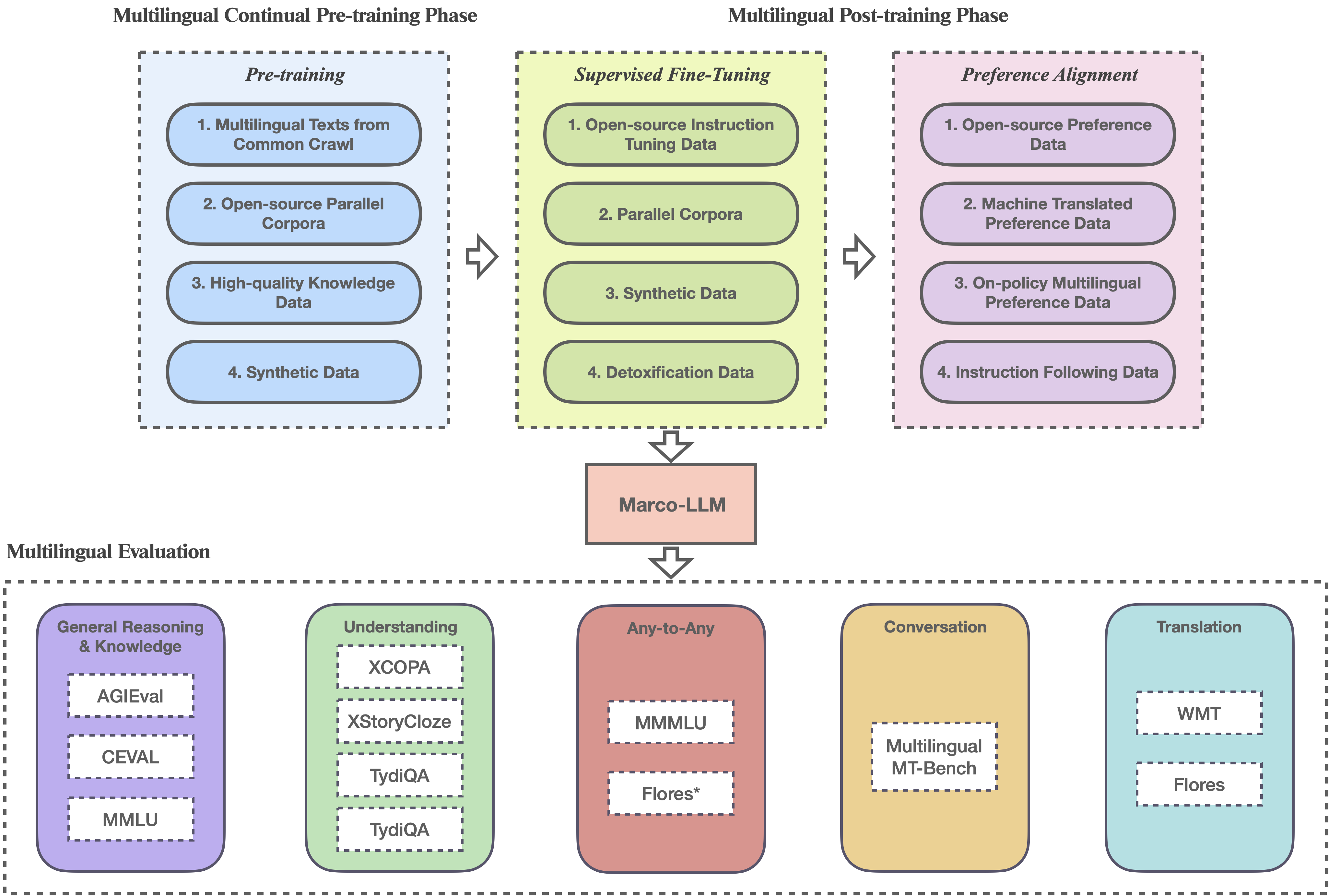}
    \caption{An overview of the training and evaluation paradigm of our Marco-LLM, we conducted massive multilingual continual pre-training, multilingual supervised finetuning and preference alignment. We further perform extensive evaluation on multilingual benchmarks to validate the efficacy of our Marco-LLM.}
    \label{fig:marco_figure}
\end{figure}

To address this disparity, we have developed \textbf{Marco-LLM}, a multilingual language model trained with a focus on low-resource languages. An overview of the framework of our Marco-LLM is shown in Figure~\ref{fig:marco_figure}. By collecting a substantial amount of multilingual data covering a series of underrepresented languages, and through massive multilingual continual pre-training and extensive multilingual post-training~(including multilingual supervised finetuning and multilingual preference alignment) using the Qwen2 model~\cite{bai2023qwen} as a foundation, Marco-LLM aims to bridge the performance gap in multilingual NLP tasks.

\noindent Our {\bf main contributions} can be summarized as follows:

\begin{itemize}[leftmargin=*,topsep=0.1em,itemsep=0.1em,parsep=0.1em]
    \item We compile and curate a large-scale multilingual dataset tailored for low-resource languages, enhancing the diversity and richness of training data.
    \item We perform massive multilingual continual pre-training and post-training on the Qwen2 model to develop Marco-LLM, a multilingual LLM that substantially improves performance on low-resource language tasks.
    \item We conduct comprehensive evaluations on benchmarks such as MMMLU, Flores, Belebele, AGIEval, multilingual MT-bench, etc, demonstrating that Marco-LLM outperforms state-of-the-art models in multilingual settings.
\end{itemize}

\noindent The rest of this paper is structured as:

\begin{itemize}[leftmargin=*,topsep=0.1em,itemsep=0.1em,parsep=0.1em]
    \item In Section~\ref{sec:related_work}, we give an overview of relevant literature regarding the development trajectory of LLMs especially multilingual LLMs and continual pre-training for LLMs.
    \item We present the details of our continual pre-training experiments in Section~\ref{sec:continual_pretraining} including the monolingual and multilingual data collection and curation, training setup and evaluation results.
    \item Furthermore, in Section~\ref{sec:post_training} we demonstrate how we conduct the post-training (including supervised finetuning and preference alignment)for the Marco-LLM that has been continual pre-trained shown in Section~\ref{sec:continual_pretraining}, including how we construct our multilingual supervised finetuning data, how to formulate the task format, supervised training details as well as corresponding evaluation results on multilingual benchmark datasets.
\end{itemize}
\section{Related Work}
\label{sec:related_work}

In recent years, Large Language Models~(LLMs) such as GPT-3~\cite{brown2020language_gpt3}, GPT-4~\cite{openai2023gpt4}, PaLM~\cite{Chowdhery2023palm}, and LLaMA~\cite{touvron2023llamaopenefficientfoundation} have revolutionized the research paradigm of Natural Language Processing~(NLP). These transformer-based~\cite{transformer} models have demonstrated exceptional capabilities in generating and understanding text, achieving state-of-the-art results in tasks like language translation, summarization, and question-answering~\cite{bang2023multitask_chatgpt_eval,wang2023document}. However, their primary focus has been on high-resource languages, leaving a gap in multilingual performance~\cite{jiao2023chatgpt,bang2023multitask_chatgpt_eval,lai2023chatgpt_English}. To bridge the gap in multilingual applications, multilingual models such as mBERT~\cite{pires2019multilingual_mbert}, XLM-R~\cite{conneau2019cross_xlm}, mT5~\cite{xue-etal-2021-mt5}, and PolyLM\cite{DBLP:journals/corr/abs-2307-06018} have been developed. These models aim to provide cross-lingual capabilities by being trained on diverse language datasets. Despite their promise, they often struggle with low-resource languages due to insufficient training data and the inherent difficulty of balancing multiple languages within one model~\cite{shi2023language_mgsm8k,dac2023okapi,singh2024indicgenbenchmultilingualbenchmarkevaluate,lovenia2024seacrowd}. Efforts in this area have included data augmentation, transfer learning, and specialized models for specific languages or tasks~\cite{conneau-etal-2018-xnli,artetxe2018unsupervised-unmt,conneau2019cross_xlm,flores101,nllb2022_flores200}. These approaches, while beneficial, have not fully harnessed the potential of large-scale language models~\cite{Ahmet2024ayamodelinstructionfinetuned}. Continual pre-training offers a viable solution to enhance the adaptability of LLMs by allowing models to incorporate new information without starting from scratch~\cite{ke2023continualpretraininglanguagemodels,yildiz2024investigatingcontinualpretraininglarg}. This method enables models to improve performance in underrepresented areas, particularly for low-resource languages. By leveraging existing strengths and optimizing computational resources, continual pre-training presents a scalable approach to overcoming current limitations in multilingual and low-resource language processing. Building on these insights, we introduce Marco-LLM, which leverages continual pre-training of the Qwen2 model with a focus on low-resource languages. Our approach integrates large-scale multilingual data collection and advanced training techniques to enhance the model's capabilities in multilingual NLP tasks.
\section{Massive Multilingual Continual Pretraining for Large Language Models}

\label{sec:continual_pretraining}

In this section, we present the technical details of how we conduct continual pretraining on LLMs. Specifically, the process of continual pretraining for multilingual LLMs encompasses the following:
(1) the curation and filtering of a large-scale training corpus, which will be introduced in Section~\ref{sec:continual_pretraining_data};
(2) simple and scalable strategies to efficiently conduct continual pretraining at large scale for LLMs, detailed in Section~\ref{sec:continual_pretraining_strategy};
(3) a comprehensive presentation of evaluation results across multiple benchmarks, along with an analysis, in Section~\ref{sec:continual_pretraining_eval_results}.

\subsection{Data Collection and Filtering}
\label{sec:continual_pretraining_data}
We create our dataset for Marco-LLM training from a variety of data sources containing knowledge until the end of 2024.4. We apply several data cleaning mechanisms and de-duplication methods on each data source to obtain high-quality tokens.

\subsubsection{Multilingual Web Data Curation}
To produce a high-quality multilingual training data, we meticulously designed a cascaded data-processing pipeline. Similar to prior processing pipelines (e.g., CCNet\cite{DBLP:conf/lrec/WenzekLCCGJG20}, RefineWeb\cite{DBLP:conf/nips/PenedoMHCACPAL23}, Llama\cite{DBLP:journals/corr/abs-2302-13971}, etc.) focusing on English,  
our multilingual pipeline features a series of data-cleaning strategies targeting text quality and information distribution. 
\begin{table}[!htb]
    \centering
    \caption{Overview of corpus utilization rates across various languages and categories of our corpus, we show the total number of tokens (in Billion Tokens) available and used for high-resource and low-resource languages, as well as other data sources such as synthetic data.}
    \label{tab:data_utilization_rate}
    \scalebox{0.75}{
    \begin{tabular}{llrrr}
        \toprule
        \textbf{Category} & \textbf{Language} & \textbf{Total Tokens (B)} & \textbf{Used Tokens (B)} & \textbf{Utilization Rate (\%)} \\ 
        \midrule
        \multicolumn{5}{l}{\textbf{High-Resource Languages}} \\
            & English (en) & 1,459.7 & 90.4 & 6.2 \\
            & Chinese (zh) & 214.7 & 48.2 & 22.4 \\
            & Arabic (ar) & 45.8 & 10.6 & 23.0 \\
            & German (de) & 442.8 & 10.6 & 2.4 \\
            & Spanish (es) & 397.8 & 10.6 & 2.7 \\
            & French (fr) & 320.8 & 10.6 & 3.3 \\
            & Korean (ko) & 41.8 & 10.6 & 25.2 \\
            & Japanese (ja) & 224.2 & 10.6 & 4.7 \\
            & Portuguese (pt) & 145.3 & 10.6 & 7.3 \\
            & Turkish (tr) & 80.6 & 10.6 & 13.1 \\ 
            \midrule
        \multicolumn{5}{l}{\textbf{Low-Resource Languages}} \\
            & Bengali (bn) & 6.5 & 1.9 & 28.7 \\
            & Hebrew (he) & 11.3 & 1.9 & 16.4 \\
            & Indonesian (id) & 23.8 & 1.9 & 7.8 \\
            & Italian (it) & 56.3 & 1.9 & 3.3 \\
            & Malay (ms) & 2.9 & 1.9 & 64.4 \\
            & Dutch (nl) & 31.3 & 1.9 & 6.0 \\
            & Polish (pl) & 45.3 & 1.9 & 4.1 \\
            & Russian (ru) & 251.0 & 1.9 & 0.7 \\
            & Thai (th) & 8.3 & 1.9 & 22.4 \\
            & Ukrainian (uk) & 18.4 & 1.9 & 10.1 \\
            & Urdu (ur) & 8.7 & 1.9 & 21.4 \\
            & Vietnamese (vi) & 24.4 & 1.9 & 7.6 \\
            & Czech (cs) & 270.2 & 1.9 & 0.7 \\
            & Greek (el) & 376.8 & 1.9 & 0.5 \\
            & Hungarian (hu) & 214.4 & 1.9 & 0.9 \\
            & Kazakh (kk) & 16.8 & 1.9 & 11.1 \\
            & Romanian (ro) & 160.0 & 1.9 & 1.2 \\
            & Azerbaijani (az) & 19.4 & 1.9 & 9.6 \\
            & Nepali (ne) & 22.6 & 1.9 & 8.2 \\ \midrule
        \multicolumn{5}{l}{\textbf{Other Data Sources}} \\
            & Parallel Data & 103.0 & 20.8 & 20.2 \\
            & High-quality Knowledge Data & 65.0 & 16.4 & 25.2 \\
            &Synthetic Data & 6.0 & 4.4 & 73.3 \\ 
            \midrule
        \textbf{Total} &  & 5,115.9 & 300.0 & 5.9 \\ 
        \bottomrule
    \end{tabular}
    }
\end{table}

\textbf{Document Preparation}. Our pipeline starts from the target document preparation to keep information distribution.
First of all, we extract the main contents from the raw Common Crawl (WARC) files. Meanwhile, we perform URL filter and language identification to avoid subsequent computationally expensive processing.
The former targets fraudulent and/or adult websites (e.g., predominantly pornographic, violent, related to gambling, etc.), while the latter focus on the target languages. Specifically, we classify documents according to their primary languages and remove those with low confidence in classification, leveraging inexpensive n-gram models (e.g, fast-Text \cite{Joulin_Grave_Bojanowski_Douze_Jégou_Mikolov_2016}).

\textbf{Quality Filtering}. The filters in this part aim for removing text with low quality. We filter out document based on some heuristic rule filters: (1). word blocklists and garbled text filters; (2). document length, the ratio of special symbols, the ratio of stop words, and the ratio of short, consecutive, or incomplete lines; (3). repeated words, n-grams. 
Inspired by CulturaX \cite{DBLP:conf/coling/NguyenNLMNDRN24}, the filtering thresholds are based on a statistical analysis of large document samples.
To enhance our filtering process,
we utilize the KenLM library \cite{DBLP:conf/lrec/WenzekLCCGJG20} to evaluate a vast array of web documents. Documents with perplexity scores largely above average are subsequently removed.

\textbf{Deduplication}. After filtering, we implement a comprehensive deduplication pipeline following the procedure in RefineWeb \cite{DBLP:conf/nips/PenedoMHCACPAL23}. This pipeline integrates document-level MinHash deduplication and sub-document exact-match deduplication, effectively identifying and removing duplicate content within and across documents. 

It's worth noting that we perform quality filtering and deduplications within data for each language. At this stage, we obtain many high-quality monolingual data in low-resource languages, generally called multilingual data, the statistics is detailed in Table \ref{tab:data_utilization_rate}. We determine the amount of multilingual tokens used in continual pretraining experimentally (shown in Section~\ref{sec:continual_pretraining_strategy}), balancing model performance on Chinese, English, and multilingual benchmarks.

\begin{figure}
    \centering
    \includegraphics[width=0.95\linewidth]{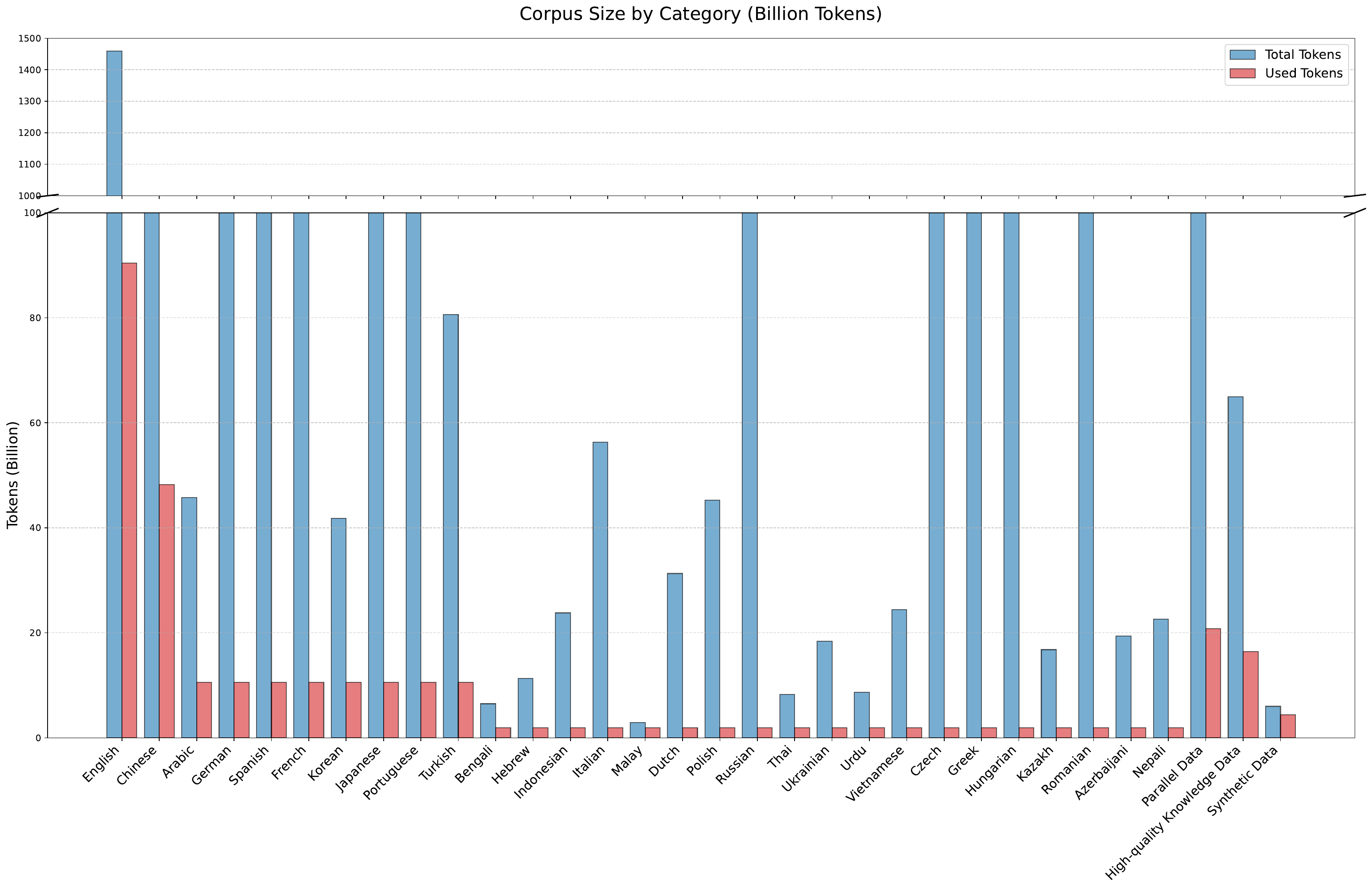}
    \caption{The amount of tokens per category in our multilingual continual pretraining corpus for Marco-LLM.}
    \label{fig:corpus_size_per_lang}
\end{figure}

\subsubsection{Parallel Data}

Follow prior works such as PaLM2 \cite{DBLP:journals/corr/abs-2305-10403}, Skywork \cite{DBLP:journals/corr/abs-2310-19341}, and PolyLM \cite{DBLP:journals/corr/abs-2307-06018}, we employ parallel data into our continual pretraining dataset to further improve the cross-lingual and multilingual ability of Marco. This data is meticulously structured to pair a complete source-language paragraph with its corresponding target-language counterpart, ensuring a seamless alignment of linguistic capabilities between the two languages. 

We mainly focus on open-source parallel data, OPUS \cite{tiedemann-2012-parallel,zhang-etal-2020-improving} and CCAligned \cite{chaudhary-EtAl:2019:WMT,elkishky_ccaligned_2020}. The former covers 100 languages, is English-centric, meaning that all training pairs include English on either the source or target side, while the latter consists of parallel or comparable web-document pairs in 137 languages aligned with English. There are many bad cases, such as translation errors, in these open-source data. We utilize the following pipeline to process them.

\textbf{Heuristic Filtering}. 
To obtain high-quality parallel corpora, we develop heuristics to remove additional low-quality documents, outliers, and documents with excessive repetitions. Some examples of heuristics include:
\begin{itemize}
\item We filter the sentence pairs using the ratio of special symbols, the ratio of stop words, the ratio of digits,, and the ratio of repeated words in source sentence; 
\item We use similarity scores of LASER embeddings \footnote{\url{https://github.com/facebookresearch/LASER}} for sentence pairs to filter out sentence pairs with low scores.
\end{itemize}

\textbf{Diverse Translation Templates}. After filtering, the translation templates are employed to concatenate the parallel corpora. Prior work \cite{DBLP:conf/acl/UstunAYKDOBSOKV24} has shown the importance of diverse wording, templates, and task types to aid generalization to different natural inputs. Therefore, we utilize diverse translation templates to make the input diversity to enhance semantic alignment between multilingual parallel sentences, the details can be found in Appendix \ref{app:translation_templates}.

Empirically, the parallel data has shown the importance of enhancing cross-lingual and multilingual ability of Marco, specific NLP downstream tasks such as machine translation. More experimental details are presented in Section \ref{sec:exp_parallel_data}.

\subsubsection{High-quality Knowledge Data}
Prior works, such as MiniCPM \cite{DBLP:journals/corr/abs-2404-06395}, Skywork \cite{DBLP:journals/corr/abs-2310-19341}, and Llama3 \cite{dubey2024llama3herdmodels}, have shown that introducing the high-quality knowledge data into the pretraining stage enhances model capabilities. Similarly, we mix some high-quality knowledge data into continual pretraining. Our high-quality knowledge data consists of mQA (multilingual Question Answer), STEM, and some ability-oriented SFT data, like math and coding. All of them are collected from the open-source website, Huggingface\footnote{\url{https://huggingface.co/}}. Similar to our pipelines for parallel multilingual data described above, we implement filters to remove data with low quality and the data from common benchmarks. In addition to the n-grams deduplications, we 
employ semantic deduplications to process them.
Note that, we do not include any training sets from commonly used benchmarks in our high-quality knowledge data.

\subsubsection{Synthetic Data}
Books and papers are high-quality knowledge data, but they are scarce.
phi models \cite{DBLP:journals/corr/abs-2306-11644, DBLP:journals/corr/abs-2309-05463} are trained by synthetic textbooks to enhance performance. Recent work \cite{DBLP:conf/emnlp/WhitehouseCA23} suggests that multilingual synthetic data can also enhance cross-lingual transfer. Also, we mix some synthetic data into our continual pretraining. Our synthetic data mainly consists of two parts: keywords-based explanation and story data, and ability-oriented SFT data. We describe them below.

\textbf{Keywords-based Explanation/Story Data}. To synthetic Wikipedia and book data, we first use GPT-4 to generate a large number of keywords covering diverse topics such as mathematics, history, physics, etc. That results in 100k keyword pools. Then, we sample several target keywords from the keyword pool and employ other LLMs with the designed prompts to generate explanations of the target words and generate textbook stories, respectively. Similar to AttrPrompt \cite{DBLP:conf/nips/YuZZMRKSZ23}, the prompts contains several attributes, we vary the value of attribute to generate diverse samples. An example of such prompts can be found in Appendix \ref{app:prompt_templates}, where the LLMs are instructed to generate training data based on attributes such as \textit{key words} and \textit{subject}.

\textbf{Ability-Oriented SFT Data}. WizardLM \cite{DBLP:conf/iclr/XuSZG0FTLJ24}, WizardMath \cite{DBLP:journals/corr/abs-2308-09583}, and WizardCoder \cite{DBLP:conf/iclr/LuoX0SGHT0LJ24} are designed to synthetic diverse instructions on target ability. We follow them to enhance Marco-LLM capacity of math and coding. To further improve the cross-lingual and multilingual ability, we use in-context learning method and translation technology to generate multilingual data. The former is that we employ the few-shot prompt to generate a new sample in the similar domain, the prompts are listed in Appendix \ref{app:prompt_templates}. The latter is that we randomly translate the original instruction, question, or answer to other language, resulting in cross-lingual QA.

To enhance the diversity of synthetic data, we employ several superb models (like GPT-4, Deepseek-v2 \cite{DBLP:journals/corr/abs-2405-04434}, DBRX\footnote{\url{ttps://www.databricks.com/blog/introducing-dbrx-new-state-art-open-llm}}, Command-R Plus \cite{cohere_for_ai_2024}, etc.) to act as the generator, as its generation is of higher quality.

\subsubsection{Data Statistics}
The composition of the continual pretraining dataset used for Marco-LLM is detailed in Table \ref{tab:data_utilization_rate}. The multilingual data mainly covers the following languages: \textit{English (en)}, \textit{Chinese (zh)}, \textit{Arabic (ar)}, \textit{German (de)}, \textit{Spanish (es)}, \textit{French (fr)}, \textit{Korean (ko)}, \textit{Japanese (ja)}, \textit{Portuguese (pt)}, \textit{Turkish (tr)}, 
\textit{Azerbaijani (az)}, \textit{Bengali (bn)}, \textit{Czech (cs)}, \textit{Greek (el)}, \textit{Hebrew (he)}, \textit{Hungarian (hu)}, \textit{Indonesian (id)}, \textit{Italian (it)}, \textit{Kazakh (kk)}, \textit{Malay (ms)}, \textit{Nepali (ne)}, \textit{Dutch (nl)}, \textit{Polish (pl)}, \textit{Romanian (ro)}, \textit{Russian (ru)}, \textit{Thai (th)}, \textit{Ukrainian (uk)}, \textit{Urdu (ur)}, \textit{Vietnamese (vi)}. The data distribution across different languages is extremely uneven, shown in Figure \ref{fig:corpus_size_per_lang}. We group them into a rough taxonomy of lower-resourced (LR) and higher-resourced (HR). This yield a split of the 29 languages in our training mixture into 10 HR and 19 LR languages. It's worth noting that our dataset contains 5.1T tokens in total, but only about 300B tokens are used to develop our Marco. We will discuss the utilization of data in Section \ref{app:data_utilization}.

\subsubsection{Data Utilization}\label{app:data_utilization}

Although we gathered a total of 5.1T training data, the actual amount used for continual pretraining is 300B. The overall data utilization rate is only 5.9\%. Table \ref{tab:data_utilization_rate} presents a detailed breakdown of the utilization of each language and data source. We have the following findings:
\begin{itemize}[leftmargin=*,topsep=0.1em,itemsep=0.1em,parsep=0.1em]
\item It is evident that, overall, the data utilization rates for both high-resource and low-resource languages are low. The more data we collect, the less effectively we utilize it. The corpus of Malay is only 2.9B, resulting in its utilization rate is up to 64.4\%. 
\item The parallel data, high-quality knowledge data, and synthetic data have higher utilization rate. These data are of high quality and focused on specific tasks; moreover, the cost of acquisition is substantial. Therefore, these data are well used. Specifically, the parallel data is used to enhance the down-stream machine translation, while the others are employed to enhance oriented ability of LLM. The quality and diversity of synthetic data enhances Marco-LLM performance, thus its utilization rate is up to 73.3\%. 
\item The multilingual capacity relies on the tokens utilized. Intuitively, to improve performance in specific languages, we should immerse the LLM in a more diverse corpus. The number of tokens used in each high-resource languages is 10.6B, compared to 1.9B in each low-resource languages.
According to Table \ref{tab:results_language_7B} and Table \ref{tab:results_language_72B}, the improvement in low-resource languages is greater than that in high-resource languages. It indicates that open-source models (e.g., Qwen2/2.5 and llama3/3.1) primarily focus on high-resource languages, with the continuous pretraining on only 1.9B corpus for each low-resource languages producing significant effects.
\end{itemize}

\noindent Based on the above findings, we outlines the following directions for further exploration:
\begin{itemize}[leftmargin=*,topsep=0.1em,itemsep=0.1em,parsep=0.1em]
\item \textbf{Further improving quality and diversity of multilingual data}. The overall data utilization rate is only 5.9\%. It means that we should further improve quality of our training data for the next model iterations. Moreover, take the diversity multilingual data into consideration, we will pay more attention to collecting multilingual data with local features.
\item \textbf{Synthetic data scaling and improving self-improvement capability}. We conclude that training with synthetic data is effective. Therefore, we will continuously focus on developing advanced techniques that can control and manipulate specific attributes of the generated data, enabling the creation of diverse and customizable synthetic datasets. In addition, we will explore methods that integrate domain-specific knowledge to ensure that the generated data adheres to the underlying constraints and patterns present in the target domain. 
Furthermore, We aim to unlock the potential of emerging self-improvement capabilities by utilizing Marco-LLM to generate synthetic data, as we believe it can potentially bootstrap its own performance by iteratively learning from the enhanced synthetic data.
\end{itemize}

\subsection{Two-Stage Continual Pretraining Strategy}
\label{sec:continual_pretraining_strategy}
We develop Marco-LLM based on Qwen2, which is naturally designed to be multilingual-friendly and has an extensive vocabulary of 150k, ensuring a high compression rate for multilingual data.
Nonetheless, the primary challenge of continual pretraining lies in balancing the adaptation (which can lead to suboptimal performance on the new dataset) and catastrophic forgetting (resulting in significant capability loss on the previous dataset). 
In this paper, we propose an advanced two-stage continual pretraining learning approach designed to facilitate the transfer of commonsense knowledge, primarily acquired in English and Chinese, to a variety of low-resource languages, as well as specific NLP downstream tasks such as machine translation.

when it comes to continual pretraining, the data mixture and the learning rate are two crucial hyper-parameters to optimize Marco. In our practice, we employ different hyper-parameters in the two-stage training. Specifically, we employ data mixing to balance the adaptation of multilingual capabilities and prevent catastrophic forgetting in Stage-I, while the goal of Stage-II is to further strengthen Marco-LLM's multilingual capabilities via a lower maximum learning rate. We describe these stages below.

\textbf{Data Mixture}.
Optimizing LLMs to learn knowledge encoded in multiple languages simultaneously is a significant challenge. We concretely formulate this problem as transferring general knowledge to low-resource languages while maintaining the advantage of original languages in the base model. To address this issue, we keep the 32\% English and 17\% Chinese corpus in Stage-I training to avoid catastrophic forgetting, as shown in Table \ref{tab:two_stage_data_mix}. Meanwhile, the proportion of other high-resourced (HR, apart from Chinese and English) and low-resourced (LR) are about 30\% and 9\%, respectively, to develop Marco-LLM with multilingual capabilities. 
The intuition is that HR has a higher priority and is more commonly used. In Stage-II, we raise a greater proportion of LR data from 9\% to 15\%, to further strength Marco-LLM multilingual capabilities in low-resourced. Moreover, parallel data, high-quality knowledge data, and synthetic data are increased, while dropping down English and Chinese corpus. Notably, the proportion of diverse corpus is determined by a large number of experiments in Marco-1.5B with a constant learning rate, not by manual experience.

\begin{table}[t]
\centering
\caption{The proportion of high-quality and multilingual sources.}
\label{tab:two_stage_data_mix}
\scalebox{0.9}{
\begin{tabular}{lrrr}
\toprule
\textbf{Data}  & \textbf{Stage-I} & \textbf{Stage-II} \\ 
\midrule
English (en)  & 32\% & 28\% \\ 
Chinese (zh) & 17\% & 15\% \\ 
High-Resourced (Others)  & 30\% & 26\% \\ 
Low-Resourced  & 9\% & 15\% \\ 
Parallel Multilingual Data & 6\% & 8\% \\ 
High-quality Knowledge Data  & 5\% & 6\% \\ 
Synthetic Data  & 1\% & 2\% \\ 
\bottomrule
\end{tabular}}
\end{table}

\textbf{Lower Maximal Learning Rate}.
Learning rate is a crucial hyper-parameter in neural network models that controls the magnitude of parameter updates \cite{DBLP:journals/tmlr/IbrahimTGRABLR24}. However, most performant open-source LLMs \cite{DBLP:journals/corr/abs-2407-10671,dubey2024llama3herdmodels} decay their learning rate to a small value (i.e. $\sim$ 1e-7) by the end of training. 
In our practice, we employ the learning rate to be re-warned and re-decayed to improve adaptation per compute spent when continual pretraining on a new distribution.
The key challenge is to optimize the maximum learning rate. 
Empirically, decreasing the schedule's maximum learning rate can help reduce forgetting, whereas increasing it can improve adaptation. We refer to Section \ref{sec:exp_learning_rate} for learning rate tuning.
After conducting fine-grained learning rate experiments when the data mixture is determined, we set maximal learning rate to $1e-5$ in Stage-I to strike a balance between the acquisition of multilingual languages and catastrophic forgetting.
Based on the HR multilingual capacities acquired from the Stage-I training, we proceed to train the LR corpus with a smaller learning rate of $6e-6$. As described in Table \ref{tab:two_stage_lr}, the training tokens in Stage-I is about 160B, while Stage-II introduces an additional 140B tokens. 
Note that, an attempt to apply the Warmup-Stable-Decay (WSD) learning rate scheduler \cite{DBLP:journals/corr/abs-2404-06395} resulted in subpar performance. It is suspected that the stable stage does not necessarily benefit model continual pretraining.

\textbf{Training}. 
Marco-LLM were trained using Pai-Megatron-LM\footnote{https://github.com/alibaba/Pai-Megatron-Patch/} on a cluster of 512 A100 GPU (64x80G) servers. To accelerate multi-node training for large model, tensor parallel and pipeline parallel were set to 8 to maximize the model flops utilization (MFU) of NVIDIA GPUs.
Specifically, we employ 512 GPU cards for Marco-72B, and 256 GPU cards for Marco-1.5B/7B.

To ensure the model learns the distribution akin, we conduct experiments in Marco-1.5B to optimize the mixing of data and learning rate. Then, we scale them up to Marco-7B and Marco-72B. 
We adopt most of the pretraining settings and model architectures from Qwen2.
During the training, all Marco-LLM were continuously pre-trained over 300B tokens, using the Adam ($\beta_1=0.9$, $\beta_2=0.95$ ) optimizer. We utilize a context window length of 32768 for Marco-1.5B and 7B, while the sequence length is 8192 in Marco-72B. At each stage, we warm-up the learning rate from 0 to the maximum learning rate over the first 10B tokens, and then decay it to 10\% of the maximal learning rate using a cosine schedule. We use a weight decay of 0.1 and gradient clipping of 1.0. 

After the two-stage continual pretraining, we have developed our multilingual large model, Marco.
Our two-stage continual pretraining is both effective and efficient to extend to the incoming rarely low-resourced languages, we leave it further exploration for future model iterations.

\begin{table}[h]
\centering
\caption{The training corpus tokens and learning rate in two Stage Continual Pretraining.}
\label{tab:two_stage_lr}
\scalebox{0.9}{
\begin{tabular}{lcc}
\toprule
\textbf{Stage} & \textbf{Training Tokens (B)} & \textbf{LR } \\ 
\midrule

Stage-I & 160 &$1e-5$ \\ 
Stage-II & 140 & $6e-6$ \\ 
\bottomrule
\end{tabular}}
\end{table}

\subsection{Evaluation Results}
\label{sec:continual_pretraining_eval_results}

 We aim to assess the capabilities of Marco-LLM from various perspectives: 1) the ability of LLMs to understand and generate natural language, as well as the ability to grasp world knowledge; 2) the performance of LLMs across different languages; and 3) their capacity to handle cross-lingual tasks such as machine translation. 
 Following the experiment design of previous work \cite{DBLP:journals/corr/abs-2307-06018}, we gather a subset of datasets from previous NLP tasks to construct a multilingual benchmark. Table \ref{tab:benchmarks} summarizes the evaluation tasks and datasets, together with their language coverage.

\begin{table}[h]
    \centering
        \caption{Evaluation benchmarks overview. The table presents the comprehensive evaluation suite used in our experiments, spanning four major task categories: general knowledge, multilingual understanding, question answering, and machine translation. Each dataset is evaluated using either accuracy (Acc.), F1 score, or BLEU metric, covering a diverse range of languages from single-language (en, zh) to multilingual scenarios.}
\scalebox{0.9}{
\begin{tabular}{llll>{\raggedright\arraybackslash}p{2.5cm}l}
    \toprule
    \textbf{Task} & \textbf{Dataset} & \textbf{Split} & \textbf{Metric} & \textbf{\#Languages} & \textbf{\#-shots} \\ 
    \midrule
    General & CEVAL & Val & Acc. & One & 5-shot \\ 
    Knowledge & AGIEval & Test & Acc. & One & 5-shot \\ 
             & ARC & Test & Acc. & One & 25-shot\\ 
             & MMLU & Test & Acc. & One & 5-shot \\ 
    \midrule
    Multilingual & XCOPA & Val & Acc. & Six & 5-shot\\ 
    Understanding & X-MMLU & Val & Acc. & Thirteen & 5-shot\\ 
                 & XStoryCloze & Val & Acc. & Six & 5-shot\\ 
    \midrule
    Question & TyDiQA & Val & F1 & Six & 1-shot\\ 
    Answering                  & Belebele & Test & Acc. & Twenty-Eight & 1-shot\\ 
    \midrule
    Machine & Flores & Devtest & BLEU & Twenty-Eight & 1-shot\\ 
    Translation & WMT-16 & Test & BLEU & Three & 1-shot\\ 
    \bottomrule
\end{tabular}
}
\label{tab:benchmarks}
\end{table}

\subsubsection{Benchmarks and Evaluation Protocol}
\label{sec_ct_experi:benchmarks_models}

All the datasets in the above multilingual benchmark can be divided into four groups: Natural Language Understanding, Knowledge, Question answering, and Machine Translation. The details of each dataset that we use for benchmarking are given below. 

\paragraph{AGIEval:} AGIEval~\cite{zhong2023agievalhumancentricbenchmarkevaluating} is a benchmark dataset designed to evaluate the reasoning and problem-solving abilities of artificial intelligence models on tasks that mimic human examinations. It includes thousands of multiple-choice questions sourced from real-world standardized tests such as the GRE, GMAT, LSAT, and other professional certification exams. The questions cover a wide range of subjects, including mathematics, logical reasoning, reading comprehension, and analytical writing. 

\paragraph{ARC (AI2 Reasoning Challenge):} The ARC dataset~\cite{clark2018thinksolvedquestionanswering_arc} consists of 7,787 natural language questions designed to assess an AI model's ability to answer grade-school-level science questions. It is divided into two subsets: \textbf{Easy Set}, containing 5,217 questions that can often be answered with surface-level information, and \textbf{Challenge Set}, comprising 2,570 more difficult questions that require reasoning and background knowledge beyond simple retrieval. Questions are multiple-choice, with four options each, covering topics like biology, physics, chemistry, and earth science. 

\paragraph{Belebele:} Belebele~\cite{bandarkar-etal-2024-belebele} is a multilingual multiple-choice reading comprehension dataset spanning 122 language variants, including low-resource and typologically diverse languages. It provides a benchmark for evaluating machine reading comprehension across a wide linguistic spectrum. Each question includes a passage, a query, and four answer choices.

\paragraph{CEVAL:} CEVAL~\cite{huang2023ceval} is a comprehensive evaluation suite designed to assess the capabilities of Chinese language models. It includes over 13,000 multiple-choice questions sourced from real Chinese national college entrance examinations and professional qualification tests. The dataset covers 52 subjects across disciplines such as mathematics, physics, law, medicine, and language arts. Each question has four options.

\paragraph{Flores:} The Flores (Facebook Low Resource Languages for Emergent Situations) dataset~\cite{nllb2022_flores200} is a multilingual machine translation benchmark focused on low-resource languages. It includes parallel sentences in over 100 languages, with a particular emphasis on underrepresented and low-resource languages. The dataset provides high-quality, professionally translated sentences, allowing for accurate evaluation of machine translation models.

\paragraph{HellaSwag:} HellaSwag~\cite{zellers2019hellaswag} is a benchmark dataset designed to test a model's ability to perform commonsense reasoning and natural language inference. It contains 70,000 multiple-choice questions generated from descriptions of everyday activities. Each question provides a context and four possible endings, with one correct continuation and three distractors that are misleading and adversarially constructed. For example: \textit{Context: "A person is cooking onions on a stove. They begin to cry because..." Options: 1. "the onions release gas that irritates the eyes." (Correct) 2. "they are listening to sad music." 3. "the stove is very hot." 4. "they forgot to buy garlic."}

\paragraph{MMLU (Massive Multitask Language Understanding):} The MMLU benchmark~\cite{DBLP:conf/iclr/HendrycksBBZMSS21_mmlu} is designed to evaluate the broad knowledge and problem-solving abilities of AI models across 57 subjects. It includes over 57,000 multiple-choice questions from high school, undergraduate, and professional levels. Subjects span various domains, including history, mathematics, law, medicine, computer science, and more. Each question has four options, and the tasks often require reasoning, calculation, or application of specialized knowledge.

\paragraph{TyDiQA:} TyDiQA (Typologically Diverse Question Answering) is a benchmark dataset~\cite{clark-etal-2020-tydiqa} for information-seeking question answering in 11 typologically diverse languages. It includes over 200,000 question-answer pairs with passages from Wikipedia in languages such as Arabic, Bengali, Finnish, Japanese, Korean, Russian, Swahili, Telugu, and others. The dataset is designed to test a model's ability to understand and generate answers in different languages without relying on English translations. TyDiQA has two primary tasks: \textbf{Gold Passage Task}, where models are provided with the correct passage containing the answer, and \textbf{Minimal Answer Task}, where models must find the minimal span in the passage that answers the question. 

\paragraph{WMT16:} The WMT16~\cite{ws-2016-machine-translation-wmt16} datasets are part of the Conference on Machine Translation's annual shared tasks, which provide standard benchmark datasets for evaluating machine translation systems. 
\textbf{WMT16} includes parallel corpora for language pairs such as English-German, English-French, English-Russian and expands based on previous years by adding languages like Romanian and incorporating more challenging test sets. 

\paragraph{XCOPA:} XCOPA~\cite{ponti-etal-2020-xcopa} is a multilingual dataset for evaluating causal commonsense reasoning in AI systems across multiple languages. It extends the original COPA (Choice of Plausible Alternatives) dataset to 11 languages, including languages like Haitian Creole, Quechua, and Yoruba. Each question consists of a premise and two alternative causes or effects, and the task is to select the more plausible one. For example: \textit{Premise: "The ground is wet." Options: 1. "It rained last night." (Cause) 2. "The sun is shining."}

\paragraph{X-MMLU:} X-MMLU~\cite{dac2023okapi} is an extension of the MMLU benchmark for evaluating multilingual models. It includes translated versions of the original MMLU tasks into multiple languages. The dataset aims to assess a model's ability to perform multitask language understanding across diverse languages, testing both its knowledge and reasoning skills in non-English contexts. X-MMLU covers subjects like mathematics, science, and humanities, requiring models to demonstrate proficiency comparable to educated human speakers in various languages.

\paragraph{XStoryCloze:} XStoryCloze~\cite{DBLP:journals/corr/abs-2112-10668_xstory_cloze} is a multilingual version of the Story Cloze Test, designed to evaluate a model's ability to understand and reason about narratives in different languages. The dataset provides short, four-sentence stories followed by two possible endings, and the task is to choose the coherent conclusion. It includes translations of the stories into multiple languages, testing narrative understanding and commonsense reasoning. For example: \textit{Story: 1. "Maria woke up early on Saturday." 2. "She was excited about the trip." 3. "She packed her bags quickly." 4. "She grabbed her keys and left the house." Endings: A. "She arrived at the airport just in time for her flight." (Correct) B. "She decided to go back to sleep because it was raining."}

\subsubsection{Baseline LLMs}

\paragraph{Llama3 and Llama3.1} The Llama 3 series includes the Llama3-8B/70B and Llama3.1-8B/70B model. These models have been pretrained on over 15 trillion tokens from publicly available sources, achieving superior performance on various benchmarks~\cite{dubey2024llama3herdmodels}.

\paragraph{Qwen2 and Qwen2.5} We compare with the Qwen2~\cite{yang2024qwen2technicalreport} series LLMs, including the Qwen2-7B/72B and Qwen2.5-7B/72B models. Qwen2 LLM is pre-trained on 7 trillion tokens and Qwen2.5 is a further optimized version of Qwen2, which is pretrained on 18 trillion tokens and achieved state-of-the-art performance on multiple evaluation benchmarks.

\subsubsection{Experimental Results}

\begin{table}[htbp]
    \centering
    \caption{Performance comparison of LLMs across various benchmarks: Results for LLMs with parameters of both 7B and 70B. The best performance in each benchmark is in bold.}
    \scalebox{0.8}{
    \begin{tabular}{lrrrrrrrrrr}
    \toprule
    \rowcolor{gray!15} \multicolumn{11}{l}{\textbf{7B Models}} \\
    Model & AGIEval & Belebele & CEval & Flores & MMLU & TyDiQA & WMT16 & XCOPA & XMMLU & XStoryCloze \\
    \midrule
    Qwen2-7B & 64.6 & 73.4 & 83.0 & 27.1 & 71.9 & 52.3 & 18.1 & 70.6 & 60.2 & 70.6 \\
    Qwen2.5-7B & 66.5 & 72.3 & 81.4 & 27.2 & \textbf{75.4} & 59.9 & 18.2 & 73.6 & \textbf{62.6} & 70.3 \\
    Llama3-8B & 24.3 & 55.3 & 37.5 & 33.1 & 53.6 & 50.5 & 24.6 & 71.7 & 49.7 & 66.5 \\
    Llama3.1-8B & 44.9 & 63.3 & 52.8 & 33.4 & 66.2 & 57.0 & 25.8 & 71.6 & 49.2 & 71.7 \\
    Marco-7B & \textbf{68.8} & \textbf{78.8} & \textbf{83.5} & \textbf{35.0} & 74.4 & \textbf{60.8} & \textbf{29.0} & \textbf{76.6} & 61.2 & \textbf{71.9} \\
    \midrule
    \rowcolor{gray!15} \multicolumn{11}{l}{\textbf{70B+ Models}} \\
    Qwen2-72B & 78.2 & 86.5 & 90.4 & 38.7 & 83.8 & 58.7 & 30.2 & 80.9 & 78.5 & 77.1 \\
    Qwen2.5-72B & 80.8 & 87.6 & 90.6 & 35.0 & {86.3} & 63.7 & 31.0 & 84.7 & 79.9 & 76.3 \\
    Llama3-70B & 60.6 & 85.5 & 66.8 & 37.4 & 79.2 & \textbf{64.3} & 34.3 & 81.1 & 72.0 & 76.9 \\
    Llama3.1-70B & 61.7 & 86.2 & 67.3 & 36.9 & 78.8 & 62.8 & 35.0 & 83.0 & 71.4 & 75.4 \\
    Marco-72B & \textbf{84.4} & \textbf{90.0} & \textbf{93.7} & \textbf{45.0} & \textbf{86.3} & 62.7 & \textbf{35.1} & \textbf{85.7} & \textbf{81.2} & \textbf{78.7} \\
    \bottomrule
    \end{tabular}
    }
    \label{tab:result_benchmark_combined_all}
\end{table}

\paragraph{Results divided by benchmarks}

Our proposed models, Marco-7B and Marco-72B, demonstrate superior multilingual capabilities across a variety of benchmarks compared to baseline models of similar sizes (see Tables~\ref{tab:result_benchmark_combined_all}). On multilingual understanding and reasoning tasks such as \textbf{Belebele}, \textbf{CEVAL}, \textbf{MMLU}, \textbf{XMMLU}, \textbf{XCOPA}, and \textbf{XStoryCloze}, the Marco-LLM consistently outperform the other strong LLMs, indicating enhanced proficiency in comprehending and common sense reasoning across diverse languages. 
For the 7B models, Marco-7B achieves the best performance across several benchmarks, obtaining the highest scores in AGIEval (68.8), Belebele (78.8), CEval (83.5), Flores (35.0), and TyDiQA (60.8). These results highlight its proficiency in handling diverse tasks and datasets, outperforming the strongest competitor, Qwen2.5-7B, by 2.3, 6.5, 2.1, 7.8, and 0.9 points, respectively. The 72B models further amplify these strengths. Marco-72B achieves the highest scores in multiple benchmarks, including AGIEval (84.4), Belebele (90.0), CEval (93.7), Flores (45.0), and XMMLU (81.2), showcasing its exceptional capacity to handle complex linguistic tasks. Compared to Qwen2.5-72B, Marco-72B surpasses by 3.6 points in AGIEval, 2.4 points in Belebele, and 10.0 points in Flores. 
Notably, Marco-72B achieves the highest scores in benchmarks like CEVAL (93.7) and XMMLU (81.4), underscoring its advanced multilingual understanding and reasoning abilities. 
In addition, the Marco-LLM exhibit strong performance in multilingual translation and generation tasks, as evidenced by competitive scores on the \textbf{Flores} and \textbf{WMT16} benchmarks. Their superior results in \textbf{TyDiQA} further highlight their capacity for multilingual question answering and knowledge retrieval. Overall, these findings supported that our focus on enhancing the multilingual abilities of LLMs has led to models that are highly effective in understanding, reasoning, and generating text across multiple languages, thereby validating the efficacy of our approach.

\begin{table}[h]
\centering
\caption{Average performance on multilingual benchmarks shown in Table~\ref{tab:benchmarks} for five 7B-parameter LLMs divided by 29 languages. The best performance for each language is highlighted in bold.}
\scalebox{0.8}{
\begin{tabular}{lrrrrr}
\toprule
\textbf{Language} & \textbf{Llama3-8B} & \textbf{Llama3.1-8B} & \textbf{Qwen2-7B} & \textbf{Qwen2.5-7B} & \textbf{Marco-7B} \\
\midrule
Chinese (zh) & 55.1 & 63.5 & 75.8 & 75.5 & \textbf{76.0} \\
English (en) & 69.6 & 74.2 & 77.4 & \textbf{78.0} & 77.9 \\
Arabic (ar) & 40.5 & 52.6 & 61.3 & 64.5 & \textbf{66.0} \\
German (de) & 47.3 & 59.8 & 69.5 & 69.0 & \textbf{72.9} \\
Spanish (es) & 57.0 & 64.9 & 70.4 & 71.9 & \textbf{72.6} \\
French (fr) & 56.0 & 63.5 & 69.8 & 71.2 & \textbf{72.5} \\
Japanese (ja) & 63.4 & 74.9 & 76.7 & 76.4 & \textbf{77.3} \\
Korean (ko) & 43.2 & 63.8 & 76.0 & \textbf{79.7} & 78.3 \\
Portuguese (pt) & 56.8 & 64.7 & 70.7 & 72.3 & \textbf{72.3} \\
Turkish (tr) & 51.1 & 62.9 & 66.0 & 66.6 & \textbf{73.4} \\
Azerbaijani (az) & 39.1 & 48.2 & 52.2 & 53.2 & \textbf{69.4} \\
Bengali (bn) & 45.9 & 49.8 & {63.9} & 62.3 & \textbf{68.9} \\
Hebrew (he) & 50.2 & 56.6 & 69.8 & 68.6 & \textbf{77.3} \\
Indonesian (id) & 64.0 & 66.1 & 77.3 & 77.8 & \textbf{82.3} \\
Italian (it) & 68.1 & 69.4 & 80.3 & 79.4 & \textbf{83.4} \\
Polish (pl) & 62.2 & 60.7 & 77.2 & 76.3 & \textbf{80.9} \\
Malay (ms) & 56.8 & 57.7 & 73.2 & 73.9 & \textbf{80.2} \\
Dutch (nl) & 61.0 & 65.2 & 80.2 & 71.6 & \textbf{82.1} \\
Romanian (ro) & 65.6 & 67.4 & 77.3 & 72.6 & \textbf{80.8} \\
Russian (ru) & 69.2 & 70.7 & 81.2 & 77.1 & \textbf{83.2} \\
Thai (th) & 54.8 & 53.3 & 69.1 & \textbf{73.7} & 72.9 \\
Ukrainian (uk) & 60.9 & 60.4 & 72.7 & 70.4 & \textbf{79.9} \\
Urdu (ur) & 50.0 & 56.7 & {63.9} & 59.9 & \textbf{71.3} \\
Vietnamese (vi) & 67.6 & 70.3 & 76.1 & {79.0} & \textbf{81.2} \\
Czech (cs) & 59.2 & 63.6 & {76.9} & 70.0 & \textbf{78.2} \\
Greek (el) & {68.1} & 67.7 & 65.0 & 67.2 & \textbf{77.1} \\
Hungarian (hu) & 59.8 & {61.0} & 63.3 & 57.9 & \textbf{69.7} \\
Kazakh (kk) & 41.3 & 44.0 & 43.1 & {45.9} & \textbf{66.1} \\
Nepali (ne) & 37.0 & 41.56 & 36.33 & {42.1} & \textbf{65.8} \\
\midrule
\textbf{Avg. Scores} & 55.9 & 61.2 & 69.4 & 69.1 & \textbf{75.5} \\
\bottomrule
\end{tabular}
}
\label{tab:results_language_7B}

\end{table}

\begin{table}[h]
\centering
\caption{Average performance on multilingual benchmarks shown in Table~\ref{tab:benchmarks} for five 70B-parameter LLMs divided by 29 languages. The best performance for each language is highlighted in bold.}
\scalebox{0.8}{
\begin{tabular}{lrrrrr}
\toprule
\textbf{Language} & \textbf{Llama3-70B} & \textbf{Llama3.1-70B} & \textbf{Qwen2-72B} & \textbf{Qwen2.5-72B} & \textbf{Marco-72B} \\
\midrule
Chinese (zh) & 75.2 & 75.0 & 83.7 & 84.8 & \textbf{86.4} \\
English (en) & 82.7 & 83.2 & 84.6 & 85.1 & \textbf{86.0} \\
Arabic (ar) & 73.0 & 73.4 & 76.2 & 77.4 & \textbf{79.7} \\
German (de) & 80.8 & 80.9 & 84.0 & 85.0 & \textbf{87.0} \\
Spanish (es) & 80.7 & 79.4 & 82.5 & 83.1 & \textbf{84.8} \\
French (fr) & 78.9 & 80.7 & 80.4 & 82.7 & \textbf{82.8} \\
Japanese (ja) & 82.5 & 84.5 & \textbf{86.6} & 86.1 & 86.2 \\
Korean (ko) & 87.1 & 87.0 & 88.8 & 88.9 & \textbf{90.6} \\
Portuguese (pt) & 81.1 & 80.8 & 83.6 & 83.8 & \textbf{85.5} \\
Turkish (tr) & 80.8 & 81.3 & 79.0 & 81.5 & \textbf{84.4} \\
Azerbaijani (az) & 77.2 & 77.9 & 78.8 & 79.1 & \textbf{85.8} \\
Bengali (bn) & 79.7 & 79.7 & 83.2 & 82.9 & \textbf{86.7} \\
Hebrew (he) & 80.1 & 82.3 & 83.9 & {84.1} & \textbf{85.1} \\
Indonesian (id) & 87.7 & 88.0 & 87.6 & 88.4 & \textbf{93.0} \\
Italian (it) & 87.1 & 87.9 & 88.1 & 89.9 & \textbf{91.0} \\
Polish (pl) & 87.2 & 87.1 & 88.6 & \textbf{88.8} & 88.2 \\
Malay (ms) & 85.2 & 87.2 & 83.4 & 87.9 & \textbf{90.7} \\
Dutch (nl) & 88.9 & 88.8 & 89.0 & 90.3 & \textbf{93.0} \\
Romanian (ro) & 88.4 & 88.2 & 88.2 & 87.3 & \textbf{90.4} \\
Russian (ru) & 87.9 & 88.3 & 89.9 & \textbf{91.0} & 90.3 \\
Thai (th) & 80.2 & 80.4 & 85.7 & 87.6 & \textbf{88.3} \\
Ukrainian (uk) & 89.0 & 88.6 & 88.2 & 90.2 & \textbf{91.7} \\
Urdu (ur) & 80.6 & 81.2 & 81.4 & 82.1 & \textbf{87.9} \\
Vietnamese (vi) & 87.1 & 89.6 & 88.7 & 90.0 & \textbf{90.8} \\
Czech (cs) & 87.1 & 87.6 & 87.8 & 90.0 & \textbf{91.4} \\
Greek (el) & 88.3 & 89.6 & 89.4 & 89.0 & \textbf{91.1} \\
Hungarian (hu) & 83.8 & 83.0 & 80.1 & {88.1} & \textbf{88.3} \\
Kazakh (kk) & 74.7 & 77.9 & 70.3 & 72.2 & \textbf{84.8} \\
Nepali (ne) & 70.1 & 73.8 & 67.1 & 74.1 & \textbf{86.7} \\
\midrule
\textbf{Avg. Scores} & 82.5 & 83.2 & 83.8 & 85.2 & \textbf{87.9} \\
\bottomrule
\end{tabular}
}
\label{tab:results_language_72B}
\end{table}

\paragraph{Results divided by languages}

The experiments evaluate the performance of our Marco-LLM: both 7B and 72B variants against various strong open-source LLMs across a diverse set of languages on the benchmarks shown in Section~\ref{sec_ct_experi:benchmarks_models}. Both Marco-LLM are built on continual pretraining on Qwen2 with massive multilingual data. The results divided by languages shown in Table~\ref{tab:results_language_7B} reveals that Marco-7B achieves a leading average score of 75.5 among the 7B parameter models. In low-resource languages such as Nepali and Kazakh, Marco-7B obtained strong performance with scores of 65.8 and 66.1, respectively, outperforming Qwen2-7B by substantial margins of 29.47 and 23.0 points. This improvement underscores the benefits of our continual pretraining approach using a vast multilingual corpus, which enhances the model's ability to generalize across languages with limited resources. Marco-7B also shows competitive performance in high-resource languages like Arabic and German, surpassing Qwen2.5-7B by 1.5 and 3.9 points, respectively. These results highlight the effectiveness of our continual pretraining approach, which improves the model's adaptability to different linguistic structures and complexities. 

Similarly, Marco-72B exhibits remarkable performance~(shown in Table~\ref{tab:results_language_72B}), achieving the highest average score of 87.9 among the LLMs with 70B+ parameters. The Marco-72B model further extends these capabilities with an average score of 87.9 across 29 languages. It demonstrates remarkable performance in low-resource languages, achieving 86.7 in Nepali and 84.8 in Kazakh, reflecting improvements over Qwen2.5-72B by 12.6 and 12.6 points, respectively. This indicates the efficacy of scaling model size alongside multilingual training data to achieve superior performance in challenging linguistic contexts. Additionally, Marco-72B remains highly competitive in high-resource languages, achieving scores of 79.7 in Arabic and 87.0 in German, which are improvements of 2.3 and 2.0 points over Qwen2.5-72B. The consistent outperformance of Marco-LLM across both high-resource and low-resource languages~(especially compared with Qwen2 - the base LLM we conduct continual pretraining, our Marco-LLM achieved substantial improvements over the languages shown in Table~\ref{tab:data_utilization_rate}) underscores the pivotal role of leveraging a vast multilingual corpus during pretraining as shown in Figure~\ref{fig:corpus_size_per_lang}. This approach not only enhances the models' ability to generalize to low-resource languages, but also maintains strong performance in high-resource languages such as English and Chinese. These findings suggest that the comprehensive training methodologies employed in the Marco-LLM are key factors contributing to their leading performance in multilingual settings.

\begin{figure}[htbp!]
    \centering
    \subfigure[]{\includegraphics[width=0.24\linewidth]{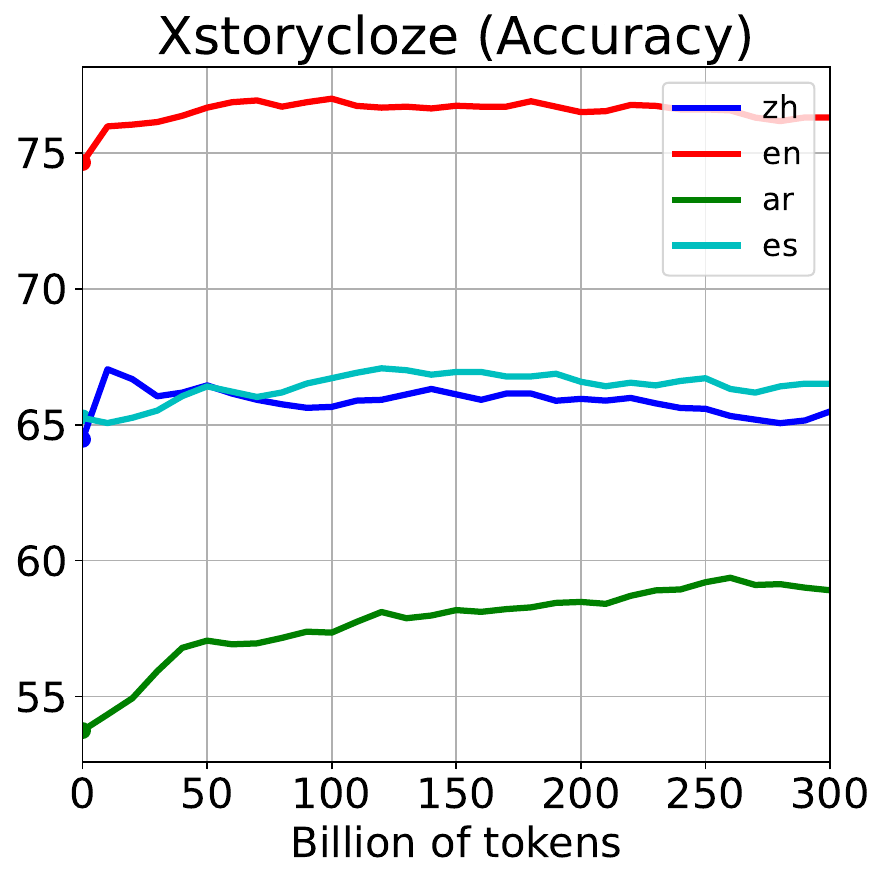}\label{exp:exp_xstorycloze}}
    \subfigure[]{\includegraphics[width=0.24\linewidth]{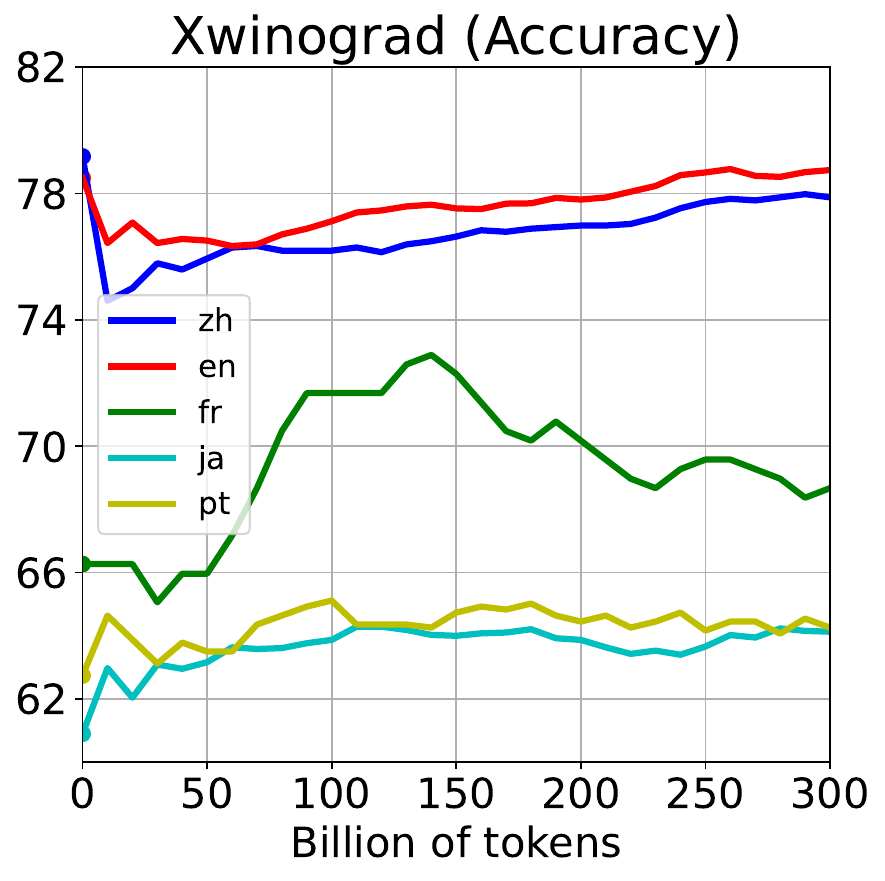}\label{exp:exp_xwinograd}}
    \subfigure[]{\includegraphics[width=0.24\linewidth]{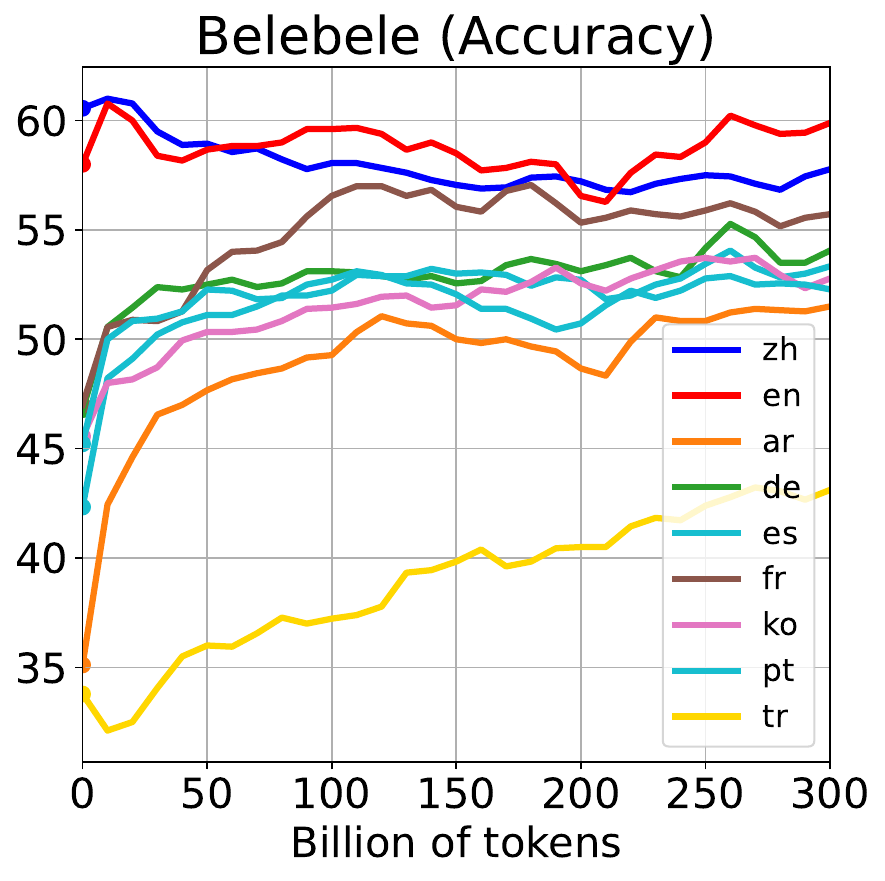}\label{exp:exp_belebele}}
    \subfigure[]{\includegraphics[width=0.24\linewidth]{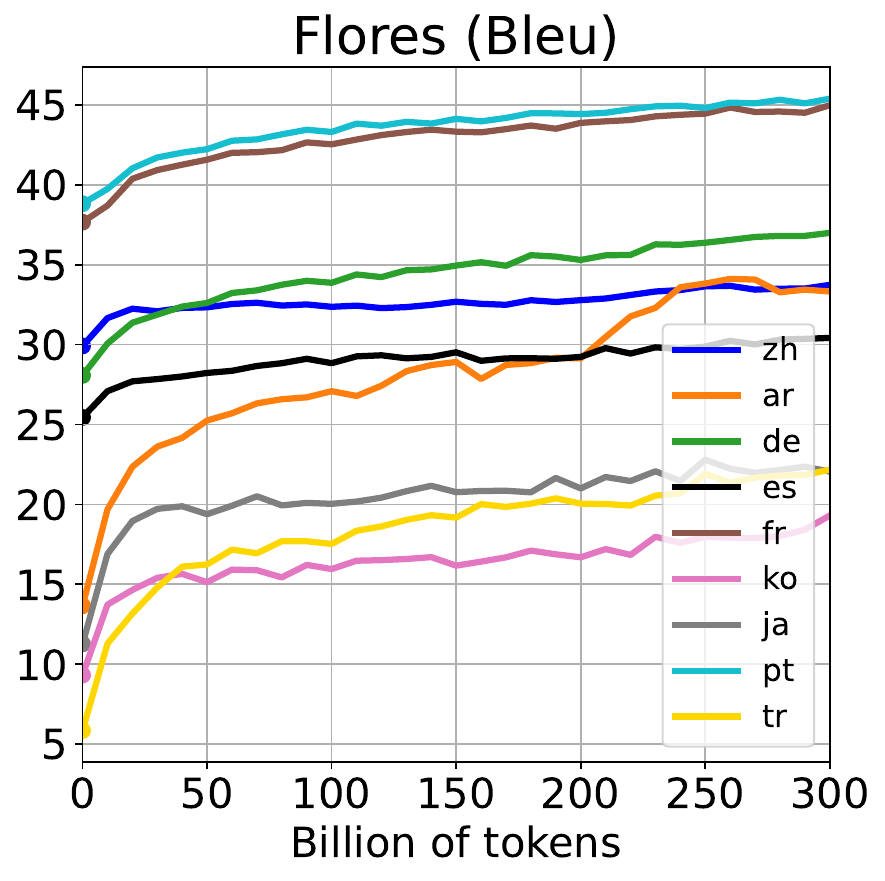}\label{exp:exp_flores}}
    \caption{Evolution of performance on question answering and machine translation during continual pretraining in Marco-1.5B.}
    \label{exp:training_dynamic}
\end{figure}

\subsection{Evolution of performance during continual pretraining}\label{sec:exp_training_dynamic}

During continual pretraining, we track the performance of our models on a few question answering and machine translation benchmarks. we report the performance of Marco-1.5B in Figure \ref{exp:training_dynamic}. 
On question answering benchmarks, the performance for multilingual languages such as \textit{Arabic (ar)}, \textit{German (de)}, \textit{Spanish (es)}, \textit{French (fr)}, \textit{Korean (ko)}, \textit{Japanese (ja)}, \textit{Portuguese (pt)}, and \textit{Turkish (tr)} improve steadily, and while capabilities in \textit{Chinese (zh)} and \textit{English (en)} are maintained, and even slightly increased. On Flores, we observe the performance of all languages shows consistent improvement. Notably, Figure \ref{exp:exp_flores} illustrates the averages for both English-to-multilingual (EN$\to$XX) and multilingual-to-English (XX$\to$EN) directions.
These indicators are consistent with our goals aiming at enhancing multilingual capabilities.

\subsection{Data ablations on Parallel Data}\label{sec:exp_parallel_data}

\begin{figure}
    \centering
    \includegraphics[width=0.45\linewidth]{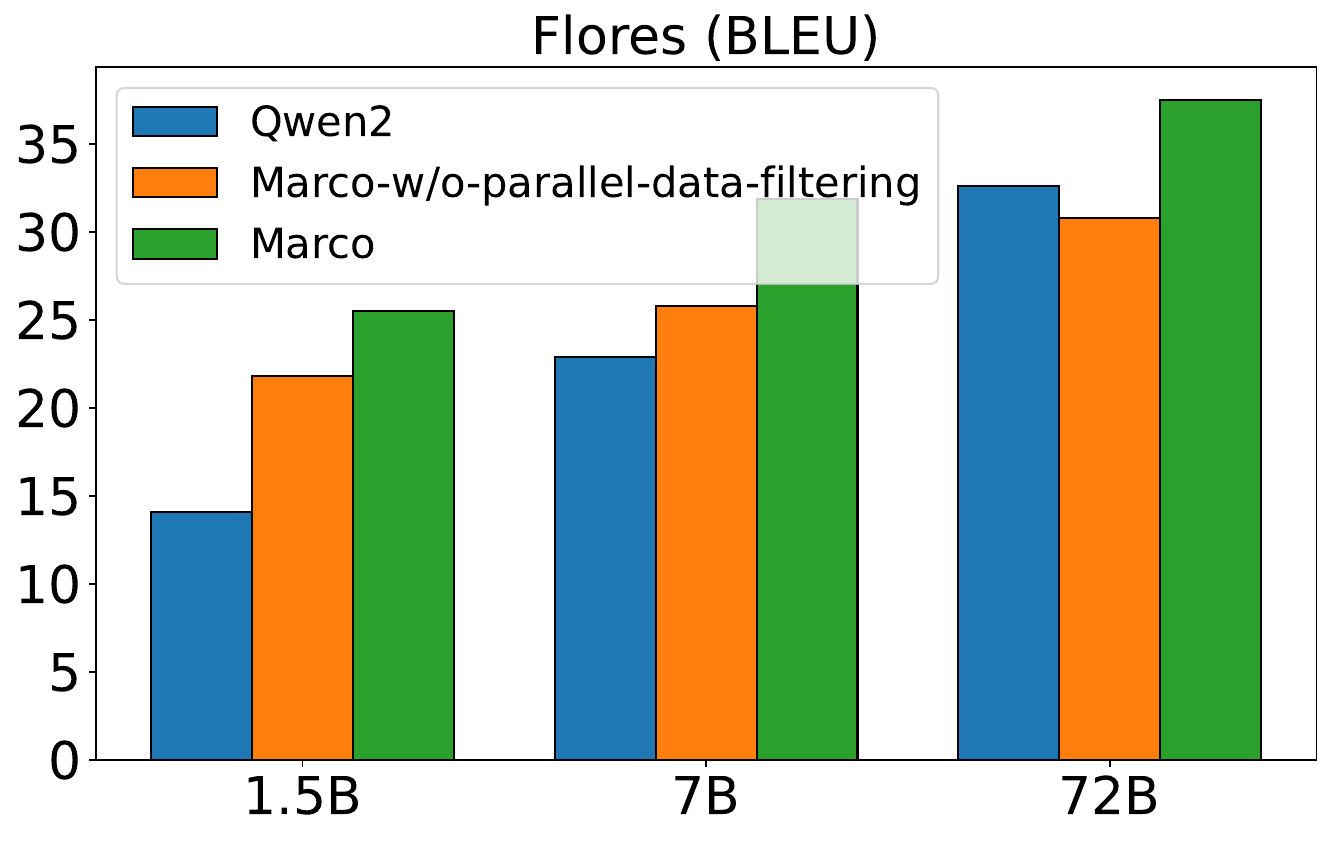}
    \caption{The performance of different model size on Flores benchmark. \textit{Marco-w/o-parallel-data-filtering} denotes that we continuously pre-trained Marco-LLM based on Qwen2 without applying any filtering to parallel data.}
    \label{fig:exp_flores_parallel_data}
\end{figure}

Empirically, introducing parallel corpora can enhance semantic alignment between cross-languages. In this section, we will explore the impact of parallel corpora on downstream specific NLP tasks, such as machine translation. We mainly focus on the quality of parallel data. The open-source parallel data has many bad cases, what happens if we ignore them on continual pretraining?
we report our findings in Figure \ref{fig:exp_flores_parallel_data}, where \textit{Marco-w/o-parallel-data-filtering} indicates that Marco-LLM were continuously pre-trained on Qwen2 without any filtering on parallel data. We have the following findings:
\begin{itemize}[leftmargin=*,topsep=0.1em,itemsep=0.1em,parsep=0.1em]
\item \textbf{Phenomena vary across different model scales}. Compared to base model Qwen2, the smaller models, specifically the 1.5B and 7B models, \textit{Marco-w/o-parallel-data-filtering} shows improvements, whereas the 72B model has performance degradation. We will elaborate on this phenomenon.
 Firstly, Marco's machine translation capability benefits from both monolingual data and parallel data, thus resulting in gradient conflicts utilizing low-quality parallel data during continual pretraining in smaller models. The monolingual data, due to its dominant proportion, plays a crucially positive role, which explains the improvements observed with \textit{Marco-w/o-parallel-data-filtering}.
 Secondly, due to parameter redundancy, the 72B model contains a higher number of monosemantic neurons\cite{DBLP:journals/tmlr/GurneeNPHTB23}. The low-quality parallel data negatively interferes with the machine translation task.  
This meaningful discovery reveals an important insights that
\textbf{some conclusions obtained on small models are not necessarily scaled to larger models}. Furthermore, beyond the issue of parallel data, we suspect that cross-language gradient conflicts may also exist during multilingual training, which we will explore in future research.

\item \textbf{High-quality data enhances the performance of model}. Compared to base model Qwen2, our Marco-LLM that has been continuously pre-trained with the processed parallel data, shows significant improvements across the 1.5B, 7B, and 72B models. We attribute it primarily to its data quality resulting from our data-engineering efforts. Furthermore, we will enhance the quality of our training data for the upcoming model iterations.

\end{itemize}

\subsection{Effect of Learning Rate}\label{sec:exp_learning_rate}

\begin{figure}[htbp!]
    \centering
    \subfigure[The average accuracy in English and Chinese with different learning rate.]{\includegraphics[width=0.36\linewidth]{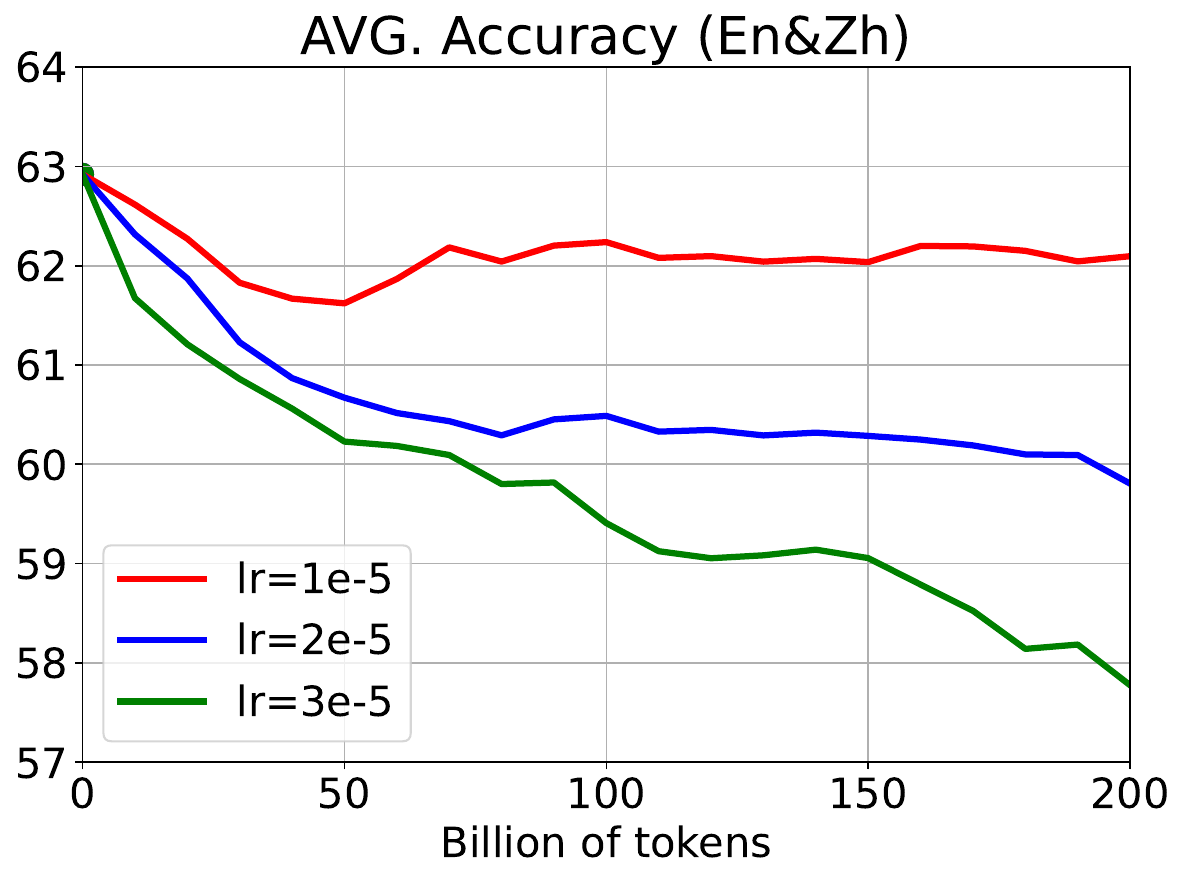}\label{fig:exp_lr_avg_en_zh}}
    \hspace{4em}
    \subfigure[The average accuracy in multilingual languages with different learning rate.]{\includegraphics[width=0.36\linewidth]{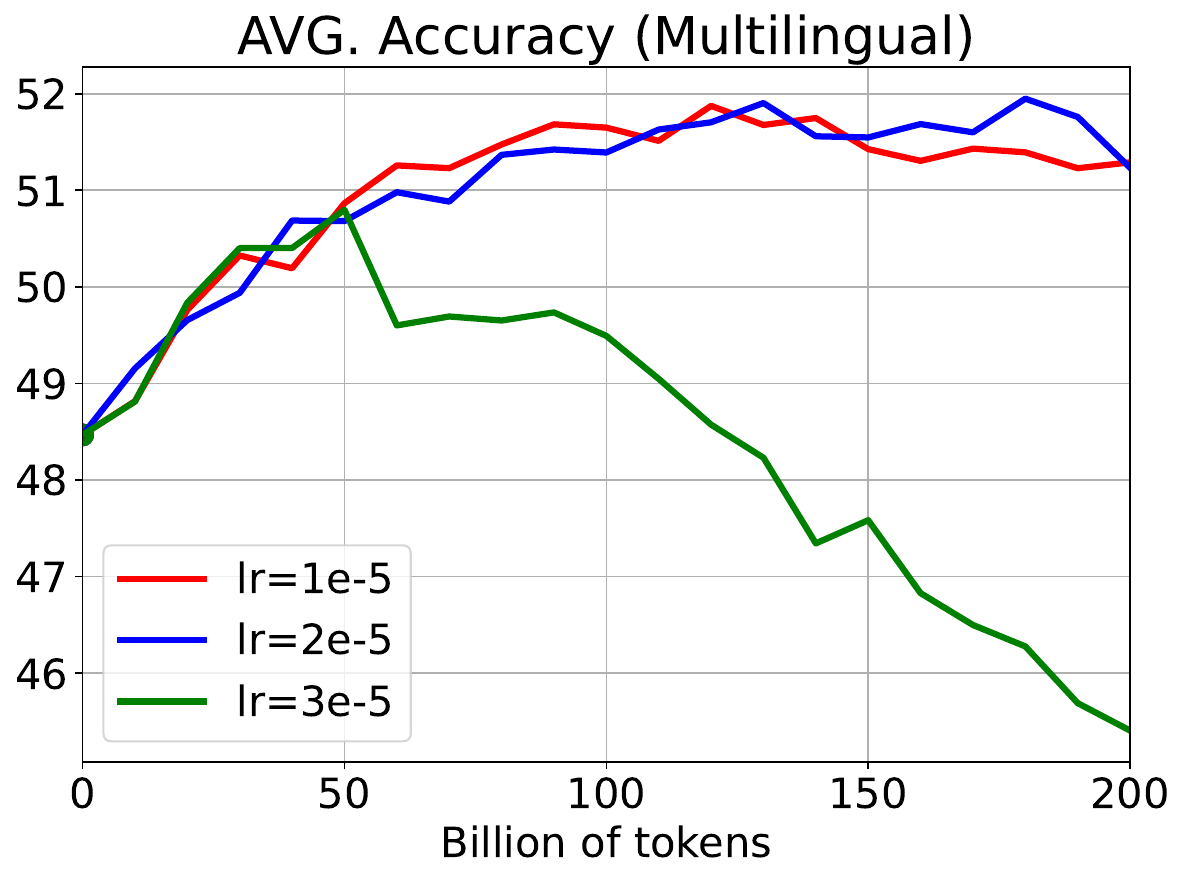}\label{fig:exp_lr_avg_multilingual}}
    \caption{The average performance on question answering benchmarks with different learning rate during continual pretraining in Marco-1.5B.}
    \label{fig:ablation_lr}
\end{figure}

In this section, we explore the effect of the crucial hyper-parameter during continual pretraining. We select learning rate (lr) from $\{1e^{-5}, 2e^{-5}, 3e^{-5}\}$ and conduct continual pretraining on 200B corpus under determined data mixture.
Figure \ref{fig:ablation_lr} demonstrates the training dynamic between the average score on question answering benchmarks and different learning rate.
Specifically, Figure \ref{fig:exp_lr_avg_en_zh} presents that the forgetting of Chinese and English ability is aggravated with the increase of learning rate, and multilingual ability is generally enhanced in Figure \ref{fig:exp_lr_avg_multilingual}. Interestingly, the average accuracy in multilingual languages goes up firstly and then down when learning rate is $3e^{-5}$. We believe that this is due to the loss of primary language ability resulting in reducing natural language understanding. Notably, the machine translation ability gets better as the learning rate increases, which shows consistently in Figure \ref{exp:exp_flores}.
Therefore, we set the peak learning rate to $1e^{-5}$ in our experiments, as it plays a pivotal role in striking a balance between the acquisition of multilingual languages and the forgetting of English and Chinese.

\section{Extensive Multilingual Post-training for Large Language Models}
\label{sec:post_training}
After conducting continuous pre-training on up to 29 languages, we proceeded with post pre-training of the Macro model. This phase of training primarily comprises two stages: Supervised Fine-Tuning (SFT)~\cite{weifinetuned,ouyangtraining-instructGPT} and Direct Preference Optimization (DPO)~\cite{rafailov2023direct_dpo}. The main objective of the Supervised Fine-Tuning (SFT) stage is to activate and enhance the model’s multilingual capabilities across various domains, including commonsense reasoning, dialogue-based question answering, precise instruction following, mathematical and logical reasoning, multilingual comprehension and translation, and coding. During this stage, our research particularly focuses on (1) the automatic generation and cost-effective collection of high-quality multilingual data, and (2) the transfer of extensive domain knowledge from high-resource languages, such as English, Chinese, and French, to low-resource languages. The DPO stage aims to ensure that the content generated by the model aligns with specified preference.

\subsection{Multilingual Supervised Fine-tuning}

\subsubsection{Data Construction}
The Supervised Fine-Tuning (SFT) dataset is composed of several components:

\begin{itemize}[leftmargin=*,topsep=0.1em,itemsep=0.1em,parsep=0.1em]
    \item A limited set of detoxification data annotated by experts, along with multilingual self-cognition enhancement data.

    \item Synthetic data and parallel corpora, which include precise instruction enhancement data, multilingual Alpaca dialogue data, as well as synthetic data for specific tasks such as comprehension, generation, and translation.

    \item Open-source instruction data, including Chain-of-Thought (CoT) enhancement data and general instruction data like Aya collection~\cite{singh2024ayadatasetopenaccesscollection}. This includes multi-turn dialogues, coding, mathematics, and more, with examples such as UltraChat, Glaive-Code, MetaMathQA, MathInstruct, Belle, and Orca.
    
    \item We utilize parallel data for multilingual machine translation tasks, enhancing language diversity and quality within the dataset. Specifically, we use the dev sets from WMT-14 to WMT-20~\cite{wmt-2020-machine} and the WikiMatrix~\cite{schwenk-etal-2021-wikimatrix}.
\end{itemize}

\paragraph{Data Collection and Processing} Our data collection endeavors encompass two primary facets. On one hand, we engage in the aggregation and cleansing of open-source data. On the other hand, we employ data synthesis and the translation of parallel corpora to augment the dataset. Building upon these two approaches, we have developed both multilingual Supervised Fine-Tuning (SFT) datasets and multilingual Direct Preference Optimization (DPO) datasets. The following sections provide a detailed examination of the composition of these datasets and the experiments conducted to assess data distribution proportions.

\textbf{Data Cleaning} Given that the majority of our dataset is derived from open-source, synthetic, and parallel corpus data, we implemented a comprehensive data processing strategy to ensure quality. Initially, we applied regular expression filtering to remove inconsistencies such as merged multi-turn dialogues, incorrect numbering in segmented outputs, HTML format outputs, emoji data, and hyperlinks or URL references. Additionally, regular expressions and Python calculations were employed to capture and validate mathematical equations, with incorrect mathematical results being discarded.

\textbf{Data Filtering} To enhance the overall performance of the model, we employed a comprehensive pipeline based on seminal works for data quality filtering, effectively removing low-quality training samples:

\begin{itemize}[leftmargin=*,topsep=0.1em,itemsep=0.1em,parsep=0.1em]
    \item Quality Scoring: For English-language data, we utilized the Deita model in conjunction to score the raw data within a range of 1 to 6. High-scoring data were selected to construct parallel corpora, with GPT-4 further filtering the translated data.
  
    \item QA similarity: In assessing QA relevance, we evaluated the semantic similarity between input and output fields. Data with similarity below a certain threshold were considered irrelevant and subsequently removed.
  
    \item Mathematics Grading: For mathematical data, we deployed models from the open-source project Open-Web-Math to score the data, retaining only those with higher scores for training purposes.
  
    \item Multilingual difficulty scoring: For extensive multilingual parallel datasets, assessing translation quality through open-source models poses challenges. We employed the Instruct-Following-Difficulty (IFD) method. By examining the ratio of Conditioned Answer Score (CA) to Direct Answer Score (DA) during the initial iterations, we filtered data that significantly benefited the model.
  
    \item Semantic deduplication: Initially, we traversed the dataset to remove duplicate instructions. We then applied MinHash and SimHash techniques for further deduplication. Lastly, embeddings were extracted using open-source models, retaining data with embedding similarity below a 0.7 threshold.
\end{itemize}

\subsubsection{Training Setup}
In our instruction fine-tuning, neat packing was employed to train a CT model with a context length of 16,384 on a our supervised finetuning data of totally 5.7 milliion examples. The training utilized the Adam optimizer with a cosine schedule learning rate. It was observed that setting a large learning rate during full model fine-tuning could severely affect the general knowledge acquired during the pre-training and CT phases, leading to a performance degradation. The optimal instruction fine-tuning learning rate was determined by adjusting the minimum pre-training learning rate in accordance with the batch size, with the maximum and minimum fine-tuning learning rates identified as 6e-6 and 6e-7, respectively.


\subsubsection{Evaluation Benchmarks}
In the evaluation experiments, we employ \textbf{TyDiQA}, \textbf{AGIEval}, \textbf{CEVAL}, \textbf{Belebele} and a multilingual version of the original MMLU:

\paragraph{Multilingual MMLU~(MMMLU)} The Multilingual Massive Multitask Language Understanding~(MMMLU) dataset\footnote{\url{https://huggingface.co/datasets/openai/MMMLU}} is an extension of the MMLU benchmark~\cite{DBLP:conf/iclr/HendrycksBBZMSS21_mmlu} into multiple languages. MMMLU includes translations of the original 57 subjects into 14 languages including Arabic, Bengali, German, Spanish, French, Hindi, Indonesian, Italian, Japanese, Korean, Brazilian Portuguese, Swahili, Yoruba, and Simplified Chinese, covering areas such as STEM, humanities, social sciences, and more. Each language version contains approximately 15,908 multiple-choice questions, mirroring the structure of the original MMLU dataset.

\subsubsection{Baseline LLMs}
\label{sec:baseline_models_post_training}
The LLMs we compared in this section are all \textit{Instruct} models by default. We employ \textbf{Llama3} and \textbf{Llama3.1} as well as \textbf{Qwen2} and \textbf{Qwen2.5}. Additionally, we also use \textbf{Aya-23} and \textbf{Aya-expanse}:

\paragraph{Aya-23 and Aya-expanse} The Aya-23 and Aya-expanse LLMs~\cite{aryabumi2024aya23openweight}~\footnote{\url{https://cohere.com/blog/aya-expanse-connecting-our-world}} that includes the Aya-23-8B, Aya-23-35B, Aya-expanse-8B and Aya-expanse-35B models, which are open-source multilingual language models supporting 23 languages. Aya-23/Aya-expanse are based on Command-R~\footnote{\url{https://cohere.com/command}} with sophisticated multilingual supervised finetuning.

\begin{table}[htbp]
    \centering
    \caption{Performance comparison of LLMs across multiple major benchmarks. Best performance in each benchmark is marked in bold.}
    \label{tab:comprehensive_results}
    \scalebox{0.8}{
    \begin{tabular}{l rrr rr}
        \toprule
        \textbf{Model} & \textbf{MMMLU} & \textbf{TydiQA} & \textbf{AGIEval} & \textbf{CEval} & \textbf{Belebele} \\
        \midrule
        \rowcolor{gray!15} \multicolumn{6}{l}{\textbf{7B Models}} \\
        Aya-23-8B       & 41.0 & 47.2 & 37.1 & 43.9 & 52.5 \\
        Aya-expanse-8B  & 48.2 & 28.3 & 36.7 & 48.5 & 64.3 \\
        Llama3-8B       & 46.6 & 39.7 & 43.4 & 50.8 & 50.7 \\
        Llama3.1-8B     & 49.2 & 53.0 & 41.8 & 55.6 & 63.9 \\
        Qwen2-7B        & 52.2 & 29.2 & 57.1 & 81.8 & 69.4 \\
        Qwen2.5-7B      & 56.0 & 39.0 & 59.0 & 77.9 & 70.0 \\
        \midrule
        Marco-Chat-7B   & \textbf{60.1} & \textbf{57.7} & \textbf{61.5} & \textbf{86.4} & \textbf{79.3} \\
        \midrule
        \rowcolor{gray!15} \multicolumn{6}{l}{\textbf{70B Models}} \\

        Aya-23-35B      & 50.1 & 50.2 & 44.4 & 53.6 & 66.3 \\
        Aya-expanse-32B & 58.9 & 30.0 & 45.7 & 56.9 & 72.7 \\
        Llama3-70B      & 64.3 & 52.0 & 57.1 & 66.7 & 76.2 \\
        Llama3.1-70B    & 71.7 & 53.1 & 55.0 & 71.6 & 84.4 \\
        Qwen2-72B       & 69.2 & 40.3 & 66.0 & 90.6 & 85.3 \\
        Qwen2.5-72B     & 69.0 & 48.4 & 67.5 & 88.2 & 88.9 \\
        \midrule
        Marco-72B       & \textbf{76.1} & \textbf{61.0} & \textbf{72.7} & \textbf{94.5} & \textbf{89.6} \\
        \bottomrule
    \end{tabular}
    }
\end{table}

\subsubsection{Results and Discussion}

\paragraph{Evaluation Results divided by benchmarks}

Table~\ref{tab:comprehensive_results} presents the average scores of our Marco-Chat-7B and Marco-72B, in comparison with several baseline models across five major benchmarks: MMMLU, TydiQA, AGIEval, CEval, and Belebele. These benchmarks are designed to evaluate language models on a diverse range of tasks and languages, highlighting their multilingual comprehension and reasoning abilities.

\begin{table}[!htb]
\centering
\caption{MMMLU Results: Performance of various LLMs across different languages in the MMMLU dataset for both 7B and 70B models. The best performance in each language is highlighted in bold.}
\label{tab:mmmlu_results_combined}
\scalebox{0.8}{
\begin{tabular}{l rrrr rrrr}
\toprule

\textbf{Language} & \textbf{Qwen2} & \textbf{Qwen2.5} & \textbf{Llama3} & \textbf{Llama3.1} & \textbf{Aya-23} & \textbf{Aya-expanse} & \textbf{Marco-Chat} & \textbf{GPT-4} \\
\midrule
        \rowcolor{lightgray} \multicolumn{9}{l}{\textbf{7B Models}} \\
Arabic & 50.9 & 56.6 & 40.5 & 42.2 & 42.1 & 48.8 & \textbf{60.6} & 62.7 \\
Bengali & 42.6 & 45.3 & 36.4 & 39.8 & 27.4 & 33.4 & \textbf{54.4} & 60.0 \\
German & 57.3 & 62.3 & 53.5 & 55.6 & 43.3 & 53.9 & \textbf{65.9} & 68.0 \\
Spanish & 60.3 & 65.3 & 55.8 & 59.1 & 47.9 & 56.1 & \textbf{67.7} & 68.2 \\
French & 61.1 & 65.0 & 55.8 & 58.9 & 46.9 & 55.5 & \textbf{67.6} & 67.5 \\
Hindi & 44.5 & 46.6 & 41.4 & 45.8 & 38.9 & 46.2 & \textbf{54.3} & 62.2 \\
Indonesian & 56.6 & 61.4 & 51.0 & 54.3 & 46.7 & 53.3 & \textbf{62.3} & 66.1 \\
Italian & 60.2 & 64.7 & 53.3 & 56.3 & 47.1 & 55.3 & \textbf{65.4} & 68.3 \\
Japanese & 56.3 & 61.0 & 42.3 & 52.1 & 45.6 & 51.5 & \textbf{64.2} & 64.4 \\
Korean & 54.1 & 59.1 & 46.5 & 50.8 & 43.6 & 50.7 & \textbf{63.0} & 63.6 \\
Chinese & 62.0 & 64.3 & 51.4 & 55.5 & 45.7 & 52.5 & \textbf{66.5} & 65.9 \\
Portuguese & 59.9 & 64.4 & 55.5 & 59.0 & 46.9 & 55.8 & \textbf{67.6} & 68.8 \\
Swahili & 34.9 & 35.7 & 37.5 & 40.3 & 26.2 & 32.0 & \textbf{43.9} & 53.1 \\
Yoruba & 30.5 & 32.8 & 31.0 & 31.4 & 26.4 & 29.9 & \textbf{37.2} & 38.0 \\
\midrule
Avg. Score & 52.2 & 56.0 & 46.6 & 50.1 & 41.0 & 48.2 & \textbf{60.1} & 62.6 \\

\midrule
        \rowcolor{lightgray} \multicolumn{9}{l}{\textbf{70B Models}} \\
Arabic & 72.0 & 74.3 & 60.6 & 71.1 & 51.8 & 61.6 & \textbf{79.3} & 71.1 \\
Bengali & 68.3 & 67.2 & 53.8 & 66.5 & 32.9 & 43.9 & \textbf{76.6} & 64.8 \\
German & 74.4 & 72.5 & 71.4 & 77.0 & 55.5 & 64.7 & \textbf{80.7} & 75.7 \\
Spanish & 77.0 & 77.5 & 74.3 & 79.3 & 58.0 & 67.5 & \textbf{82.6} & 76.8 \\
French & 75.6 & 76.0 & 73.1 & 77.9 & 58.1 & 67.5 & \textbf{80.7} & 75.8 \\
Hindi & 69.9 & 69.1 & 65.0 & 72.7 & 47.6 & 58.8 & \textbf{76.9} & 70.1 \\
Indonesian & 73.1 & 73.3 & 70.6 & 75.7 & 55.5 & 65.4 & \textbf{79.0} & 73.7 \\
Italian & 75.3 & 72.5 & 73.3 & 77.8 & 57.8 & 66.5 & \textbf{81.6} & 75.8 \\
Japanese & 74.1 & 74.7 & 65.6 & 73.8 & 54.5 & 64.3 & \textbf{81.6} & 71.6 \\
Korean & 72.3 & 71.8 & 64.5 & 72.7 & 53.7 & 62.4 & \textbf{78.8} & 71.3 \\
Chinese & 77.5 & 76.7 & 69.5 & 74.8 & 54.1 & 63.4 & \textbf{82.0} & 72.5 \\
Portuguese & 76.8 & 76.9 & 73.7 & 78.9 & 58.3 & 67.2 & \textbf{81.7} & 76.2 \\
Swahili & 47.3 & 48.8 & 51.1 & 64.0 & 33.6 & 38.4 & \textbf{63.7} & 68.1 \\
Yoruba & 34.6 & 35.5 & 33.6 & 41.2 & 30.4 & 33.4 & \textbf{44.0} & 47.3 \\
\midrule
Avg. Score & 69.2 & 69.0 & 64.3 & 71.7 & 50.1 & 58.9 & \textbf{76.1} & 70.8 \\
\bottomrule
\end{tabular}
}
\end{table}

Our Marco-Chat-7B model consistently achieves the highest scores among the 7B parameter models across all benchmarks. Specifically, it significantly outperforms the baselines on CEval and Belebele, which focus on Chinese educational subjects and a variety of African languages, respectively. On CEval, Marco-Chat-7B attains a score of 86.4, surpassing the next best model, Qwen2-7B, by a substantial margin of 4.6 points. This indicates our model's strong capability in understanding and processing Chinese language content. Similarly, on Belebele, which evaluates proficiency in underrepresented African languages, Marco-Chat-7B achieves a score of 79.3, outperforming others by nearly 9 points. This demonstrates the effectiveness of our multilingual training approach in capturing linguistic nuances across diverse languages, including those with limited available data. In the 70B size LLMs, Marco-72B leads the group by a large margin on all benchmarks. It achieves a score of 76.1 on MMMLU, which assesses multitask language understanding across various subjects and languages. This result is 4.4 points higher than the next best model, Llama3.1-70B, highlighting our model's superior general language understanding capabilities. On TydiQA, a benchmark for typologically diverse question answering, Marco-72B attains a score of 61.0, outperforming the second-best model by 7.9 points. This suggests that our model shows strong performance at various tasks across a wide range of languages with different grammatical structures and scripts. Additionally, Marco-72B achieves an impressive score of 94.5 on CEval, indicating excellent proficiency in Chinese across various academic subjects. On Belebele, it reaches 89.6, showcasing strong performance in languages that are often underrepresented in training data.

These results highlight the effectiveness of our approach in building a truly multilingual LLM. By consistently achieving top performance across benchmarks that evaluate different languages and tasks, our models demonstrate robust multilingual proficiency and adaptability. This underscores the importance of incorporating a diverse and comprehensive multilingual dataset during training to enhance the language understanding abilities of large language models. Our observations also reveal that models with a higher number of parameters, like Marco-72B, substantially benefit from our multilingual training strategy, achieving significant improvements over other large models. Furthermore, the consistent gains across both high-resource languages (like English and Chinese) and low-resource languages (as represented in Belebele) indicate that our models do not merely rely on the abundance of data in certain languages but truly learn to generalize across linguistic boundaries.

\begin{table}[t!]
\centering
\caption{Performance comparison of language models on Belebele benchmark~\cite{bandarkar-etal-2024-belebele} across different languages.}
\label{tab:belebele_results}
\scalebox{0.8}{
\begin{tabular}{l rrrr rrr}
\toprule

\textbf{Language} & \textbf{Qwen2} & \textbf{Qwen2.5} & \textbf{Llama-3} & \textbf{Llama3.1} & \textbf{Aya-23} & \textbf{Aya-expanse} & \textbf{Marco-Chat} \\
\midrule
        \rowcolor{lightgray} \multicolumn{8}{l}{\textbf{7B Models}} \\
Azerbaijani & 59.9 & 60.4 & 41.3 & 57.7 & 34.1 & 49.9 & \textbf{72.3} \\
Bengali & 64.9 & 64.2 & 46.8 & 57.2 & 28.7 & 41.6 & \textbf{75.1} \\
Czech & 75.4 & 75.9 & 54.3 & 73.4 & 61.6 & 76.9 & \textbf{84.4} \\
Greek & 68.7 & 75.2 & 59.0 & 74.3 & 65.2 & 80.3 & \textbf{81.4} \\
Hebrew & 74.2 & 72.7 & 45.9 & 59.3 & 61.7 & 77.9 & \textbf{82.1} \\
Hungarian & 57.3 & 63.0 & 45.1 & 52.4 & 35.0 & 44.0 & \textbf{68.0} \\
Indonesian & 77.2 & 78.4 & 61.9 & 75.0 & 64.7 & 77.6 & \textbf{82.8} \\
Italian & 78.2 & 81.9 & 60.6 & 71.8 & 67.0 & 75.7 & \textbf{86.8} \\
Japanese & 78.0 & 69.8 & 49.9 & 64.7 & 64.2 & 74.1 & \textbf{83.1} \\
Kazakh & 48.0 & 51.2 & 36.0 & 49.1 & 28.3 & 38.0 & \textbf{73.1} \\
Malay & 78.9 & 77.1 & 57.9 & 67.4 & 53.2 & 73.3 & \textbf{83.4} \\
Dutch & 58.7 & 74.6 & 56.3 & 70.1 & 66.2 & 72.1 & \textbf{85.3} \\
Nepali & 44.3 & 49.4 & 38.9 & 46.7 & 32.9 & 40.8 & \textbf{70.0} \\
Polish & 78.4 & 70.3 & 55.0 & 65.0 & 61.1 & 73.9 & \textbf{73.3} \\
Romanian & 72.4 & 74.0 & 55.2 & 68.7 & 65.9 & \textbf{74.8} & 73.2 \\
Russian & 79.7 & 73.3 & 52.7 & 70.2 & 64.1 & 74.3 & \textbf{87.9} \\
Ukrainian & 76.1 & 73.2 & 51.1 & 69.8 & 61.8 & 76.2 & \textbf{83.4} \\
Urdu & 64.9 & 64.4 & 49.0 & 59.2 & 35.6 & 50.2 & \textbf{76.2} \\
Thai & 72.3 & 74.5 & 43.0 & 57.2 & 37.6 & 41.8 & \textbf{76.9} \\
Vietnamese & 80.0 & 77.0 & 53.8 & 68.3 & 61.4 & 73.2 & \textbf{86.3} \\
\midrule
Avg. Scores & 69.4 & 70.0 & 50.7 & 63.9 & 52.5 & 64.3 & \textbf{79.3} \\
\midrule
        \rowcolor{lightgray} \multicolumn{8}{l}{\textbf{70B Models}} \\
Azerbaijani & 79.9 & 81.6 & 63.3 & 79.7 & 51.9 & 58.3 & \textbf{85.6} \\
Bengali & 84.9 & 87.3 & 75.3 & 81.4 & 42.1 & 64.8 & \textbf{89.2} \\
Czech & 89.7 & 91.9 & 79.0 & 86.8 & 79.6 & 85.8 & \textbf{91.8} \\
Greek & 89.2 & \textbf{92.6} & 87.0 & 89.4 & 80.1 & 83.6 & 91.9 \\
Hebrew & 85.0 & 86.9 & 75.9 & 78.9 & 77.0 & 83.1 & \textbf{86.0} \\
Hungarian & 72.8 & \textbf{89.3} & 58.0 & 74.1 & 53.6 & 50.8 & 87.0 \\
Indonesian & 88.9 & 91.7 & 82.6 & 87.3 & 77.8 & 81.1 & \textbf{93.1} \\
Italian & 89.8 & 90.4 & 83.8 & 87.9 & 81.1 & 77.7 & \textbf{91.1} \\
Japanese & 87.8 & 90.2 & 82.2 & 86.9 & 73.7 & 78.2 & \textbf{90.1} \\
Kazakh & 73.6 & 76.0 & 54.1 & 78.2 & 40.7 & 55.1 & \textbf{81.7} \\
Malay & 88.7 & 91.2 & 87.7 & 88.7 & 74.8 & 76.4 & \textbf{92.1} \\
Dutch & 90.9 & 93.2 & 80.4 & 87.1 & 80.8 & 85.7 & \textbf{94.4} \\
Nepali & 70.6 & 80.1 & 55.7 & 76.0 & 39.9 & 50.1 & \textbf{84.4} \\
Polish & 86.4 & 89.7 & 75.0 & 88.7 & 73.9 & 83.7 & \textbf{90.6} \\
Romanian & 88.4 & 92.1 & 73.4 & 86.3 & 75.7 & 77.7 & \textbf{90.6} \\
Russian & 90.0 & 94.1 & 86.1 & 87.2 & 73.2 & 84.6 & \textbf{92.7} \\
Ukrainian & 90.9 & \textbf{93.4} & 84.1 & 90.2 & 74.3 & 79.8 & 93.0 \\
Urdu & 83.1 & 86.4 & 78.9 & \textbf{90.2} & 47.2 & 63.7 & 88.2 \\
Thai & 86.2 & 87.1 & 79.7 & 82.7 & 53.6 & 63.8 & \textbf{87.6} \\
Vietnamese & 88.8 & 92.0 & 81.7 & 83.9 & 75.8 & 69.2 & \textbf{92.2} \\
\midrule
Avg. Scores & 85.3 & 88.9 & 76.2 & 84.4 & 66.3 & 72.7 & \textbf{89.6} \\
\bottomrule
\end{tabular}
}
\end{table}

\begin{table}[t!]
    \centering
    \caption{Performance comparison of various LLMs and MT systems on the Flores benchmark. The best performance in each column is highlighted in bold.}
    \label{tab:mt_flores}
    \scalebox{0.63}{
    \begin{tabular}{l rrrrr rrrrr r}
        \hline
        \toprule
        & \textbf{GPT4} & \textbf{DeepL} & \textbf{Google} & \textbf{Aya-32B} & \textbf{Aya-35B} & \textbf{Qwen2-72B} & \textbf{Qwen2.5-72B} & \textbf{Llama3-70B} & \textbf{Llama3.1-70B} & \textbf{Marco-7B} & \textbf{Marco-72B}  \\
        \midrule

        \rowcolor{lightgray} En$\to$XX & &  & & & & & & & & & \\

En2Ar &  40.4  &  48.1  &  50.0  &  31.5 &  24.4 &  17.1 &  29.9 &  29.2 &  33.5 &  41.5  &  \textbf{61.2} \\  \\
En2De &  45.9  &  48.7  &  \textbf{49.3}  &  32.3 &  33.1 &  37.7 &  40.4 &  43.0 &  44.7 &  41.6  &  47.7 \\  \\
En2Es &  33.1  &  32.9  &  34.6  &  28.4 &  20.6 &  32.0 &  31.7 &  31.5 &  32.5 &  33.1  &  \textbf{37.2} \\  \\
En2Fr &  54.4  &  59.1  &  57.9  &  50.5 &  34.5 &  52.3 &  53.2 &  52.6 &  55.2 &  54.9  &  \textbf{58.8} \\  \\
En2It &  37.2  &  41.5  &  39.1  &  32.4 &  25.3 &  34.2 &  34.8 &  34.8 &  36.6 &  36.2  &  \textbf{40.6} \\  \\
En2Ja &  34.6  &  36.8  &  \textbf{41.1}  &  31.0 &  9.6 &  29.6 &  33.0 &  14.3 &  33.0 &  36.9  &  37.7 \\  \\
En2Ko &  28.5  &  32.9  &  \textbf{33.7}  &  27.2 &  12.3 &  19.2 &  24.8 &  0.1 &  27.7 &  29.0  &  31.6 \\  \\
En2Nl &  34.8  &  37.0 &  \textbf{36.3}  &  25.4 &  24.2 &  28.4 &  30.9 &  32.0 &  34.7 &  33.0  &  35.9 \\  \\
En2Pl &  30.3  &  33.4  &  \textbf{33.7}  &  24.8 &  16.5 &  20.4 &  26.0 &  28.6 &  29.5 &  28.3  &  30.6 \\  \\
En2Pt &  54.8  &  45.7  &  56  &  44.0 &  41.7 &  50.1 &  52.6 &  52.0 &  54.8 &  54.5  &  \textbf{57.9} \\  \\
En2Ru &  36.8  &  40.5  &  40.9  &  32.3 &  23.8 &  33.6 &  36.1 &  35.6 &  37.9 &  37.7  &  \textbf{43.4} \\  \\
En2Tr &  36.9  &  \textbf{45.0}  &  44.2  &  33.1 &  26.4 &  22.8 &  30.4 &  32.7 &  36.8 &  35.8  &  37.7 \\  \\
En2Uk &  37.0 &  42.9  &  41.6  &  33.5 &  17.0 &  25.2 &  30.0 &  36.5 &  36.8 &  36.3  &  \textbf{44.3} \\  \\
En2Zh &  44.2  &  48.6  &  \textbf{50.6}  &  26.0 &  15.6 &  28.0 &  33.9 &  13.3 &  31.3 &  45.3  &  48.6 \\  \\
        \midrule
        Avg. Scores& 39.2 & 42.4 & 43.5 & 32.3 &  23.2 &  30.8 &  34.8 &  31.2 &  37.5 & 38.9 &  \textbf{43.8} \\ 
        \midrule
        \rowcolor{lightgray} XX$\to$En & &  & & & & & & & & & \\

Ar2En &  42.7  &  47.7  &  46.8  &  33.3 &  41.1 &  41.5 &  44.6 &  37.1 &  46.1 &  48.0  &  \textbf{58.0} \\  \\
De2En &  47.7  &  51.0  &  51.3  &  30.8 &  40.9 &  46.9 &  48.4 &  46.3 &  49.4 &  50.6  &  \textbf{54.7} \\  \\
Es2En &  34.3  &  36.9  &  36.3  &  24.8 &  33.8 &  34.5 &  35.3 &  33.7 &  35.0 &  40.2  &  \textbf{47.2} \\  \\
Fr2En &  48.9  &  50.8  &  52.7  &  27.7 &  45.1 &  48.8 &  49.5 &  47.5 &  50.7 &  51.5  &  \textbf{56.8} \\  \\
It2En &  36.7  &  40.2  &  40.2  &  28.6 &  37.5 &  36.7 &  38.2 &  36.5 &  38.4 &  42.9  &  \textbf{49.2} \\  \\
Ja2En &  30.4  &  37.0  &  36.7  &  20.5 &  22.6 &  29.8 &  31.9 &  26.2 &  32.3 &  36.3  &  \textbf{49.5} \\  \\
Ko2En &  33.3  &  39.3  &  38.2  &  21.9 &  25.9 &  32.1 &  34.5 &  28.7 &  33.8 &  37.0  &  \textbf{49.0} \\  \\
Nl2En &  36.0  &  37.7  &  38.7  &  23.3 &  32.8 &  35.9 &  36.3 &  34.8 &  37.0 &  39.8  &  \textbf{46.4} \\  \\
Pl2En &  33.5  &  35.8  &  37.0  &  19.6 &  27.6 &  33.7 &  34.7 &  32.1 &  35.3 &  38.4  &  \textbf{45.9} \\  \\
Pt2En &  53.1  &  55.8  &  56.3  &  37.9 &  50.7 &  53.0 &  53.5 &  51.8 &  54.9 &  54.5  &  \textbf{60.3} \\  \\
Ru2En &  38.7  &  43.3  &  42.9  &  23.6 &  36.2 &  39.0 &  40.2 &  37.9 &  41.0 &  43.8  &  \textbf{49.2} \\  \\
Tr2En &  42.6  &  48.5  &  47.7  &  31.1 &  36.5 &  39.9 &  42.3 &  37.9 &  43.1 &  43.4  &  \textbf{52.0} \\  \\
Uk2En &  43.4  &  47.2  &  47.3  &  27.4 &  38.8 &  41.6 &  43.7 &  40.5 &  44.9 &  46.3  &  \textbf{58.8} \\  \\
Zh2En &  31.3  &  36.8  &  37.7  &  24.3 &  26.7 &  31.0 &  35.4 &  29.0 &  34.7 &  38.2  &  \textbf{45.5} \\  \\ 
        \midrule
        Avg. Scores& 36.8 & 43.4 & 43.6 & 26.8 &  35.4 &  38.9 &  40.6 &  37.2 & 41.2 & 43.6 &  \textbf{51.6} \\ 
        \bottomrule
    \end{tabular}
    }

\end{table}

\paragraph{MMMLU Results}

We also highlight the results on the recently released multilingual version of MMLU from OpenAI~\footnote{\url{https://huggingface.co/datasets/openai/MMMLU}}, which are shown in Table~\ref{tab:mmmlu_results_combined}. The MMMLU dataset evaluates language models across a diverse set of languages and tasks. This benchmark is crucial for assessing the multilingual capabilities of language models, particularly their ability to process and understand languages beyond English.

In the 7B LLMs, Marco-7B consistently outperforms baseline models across a wide range of languages. It achieves the highest scores in languages such as Arabic (60.6), Bengali (54.4), German (65.9), and Spanish (67.7), demonstrating a significant advantage over other models. This performance is particularly noteworthy in Bengali, where Marco-7B surpasses the second best model~(Qwen-2.5-7B) by a substantial margin of 4.1, highlighting its ability to handle languages with less training data effectively. The model's strong results in languages like Hindi~(+7.7 compared to the second best model) and Korean further emphasize its robust multilingual capabilities, making it a versatile tool for diverse linguistic contexts. In the 70B LLMs, Marco-72B achieves the highest average score across all languages, showcasing its superior multilingual understanding. It demonstrated competitive performance in high-resource languages such as German (80.7), Spanish (82.6), and Chinese (82.0), while also delivering strong performance in lower-resource languages like Swahili (63.7) and Yoruba (44.0). These results underscore the model's extensive language processing abilities, positioning it as a leading solution for multilingual applications. It is worth noting that our Marco-72B outperformed GPT-4~(for 7B LLMs we compare with GPT-4o-mini and for 70B LLMs we compare GPT-4)~\cite{hurst2024gpt4o} in many languages.
truction (existing preference dataset)

\paragraph{Belebele Results}

The results on the Belebele ~\cite{bandarkar-etal-2024-belebele}, presented in Table~\ref{tab:belebele_results}, illustrate the strong multilingual capabilities of the Marco-Chat models across both 7B and 70B parameter scales. 

For the 7B models, Marco-Chat consistently achieves the highest scores across most languages, with an average score of 79.3, significantly outperforming other models such as Qwen2 (69.4) and Qwen2.5 (70.0). This performance is particularly impressive in low-resource languages like Kazakh and Nepali, where Marco-Chat scores 73.1 and 70.0, respectively, showcasing improvements of 25.1 and 25.7 points over Qwen2. These substantial gains underscore the success of our continual pretraining approach, which effectively leverages a vast multilingual corpus to enhance generalization capabilities across languages with limited resources. Additionally, Marco-Chat remains highly competitive in high-resource languages such as Italian (86.8) and Japanese (83.1), further demonstrating its versatility and robustness. The 70B models exhibit similar patterns, with Marco-Chat achieving an average score of 89.6, leading the performance across nearly all languages. Notable improvements are observed in Bengali (89.2) and Indonesian (93.1), surpassing Qwen2.5 by 1.9 and 1.4 points, respectively. The model's ability to handle complex linguistic tasks is further exemplified in high-resource languages such as Dutch (94.4) and Russian (92.7), where it achieves top scores. These results highlight the strategic advantage of our approach, which effectively captures linguistic nuances and complexities, benefiting from extensive multilingual pretraining. Overall, the results on the Belebele benchmark validate our focus on enhancing multilingual capabilities. 

\begin{table}[t!]
\centering
\label{tab:mt_any2any}
\caption{\textit{Any2Any} translation performance on Flores benchmark across various language pairs for LLMs. The table compares results from the \textit{Instruct} models of Qwen2, Qwen2.5, Llama3, Llama3.1, and Marco-LLM, with the best performance highlighted in bold for each translation direction.}
\scalebox{0.8}{
\begin{tabular}{lrrrrrrr}
\toprule
Trans. Dir. & Qwen2-7B & Qwen2.5-7B & Llama3-8B & Llama3.1-8B & Aya-expanse-8B  & Aya-23-8B & Marco-7B\\
\midrule
Ar2Ja & 16.2 & 14.3 & 13.1 & 17.5 & 16.9 & 17.1 & \textbf{21.8} \\
Es2Ja & 17.2 & 10.7 & 11.2 & 18.1 & 18.3 & 19.9 & \textbf{22.4} \\
Fr2Ja & 19.9 & 14.0 & 14.1 & 21.6 & 17.6 & 22.3 & \textbf{25.8} \\
Hu2Ja & 15.2 & 9.6 & 12.3 & 16.0 & 10.7 & 12.6 & \textbf{20.3} \\
Hu2Ko & 10.5 & 6.9 & 9.4 & 10.2 & 9.2 & 11.0 & \textbf{14.0} \\
Ja2Ar & 9.8 & 7.4 & 8.6 & 11.1 & 12.3 & 9.6 & \textbf{15.5} \\
Ja2Es & 16.1 & 15.0 & 15.9 & 16.7 & 12.9 & 16.2 & \textbf{19.1} \\
Ja2Ko & 16.3 & 12.1 & 12.2 & 17.3 & 18.7 & 17.2 & \textbf{22.6} \\
Ja2Th & 11.7 & 11.4 & 11.1 & 11.6 & 0.8 & 2.0 & \textbf{16.4} \\
Ja2Zh & 19.5 & 17.6 & 6.9 & 10.2 & 14.9 & 12.7 & \textbf{22.9} \\
Kk2Ar & 7.3 & 5.5 & 7.1 & 8.6 & 2.3 & 5.7 & \textbf{11.6} \\
Kk2Fr & 13.8 & 8.9 & 17.9 & 14.1 & 5.6 & 10.6 & \textbf{20.5} \\
Kk2Ja & 11.7 & 6.0 & 10.4 & 12.2 & 4.1 & 9.6 & \textbf{16.8} \\
Kk2Ko & 8.3 & 4.7 & 9.6 & 9.3 & 4.5 & 7.5 & \textbf{13.1} \\
Kk2Pt & 12.7 & 8.8 & 15.2 & 10.2 & 4.2 & 9.7 & \textbf{16.9} \\
Kk2Th & 6.7 & 7.1 & 8.9 & 10.4 & 0.4 & 1.2 & \textbf{12.7} \\
Kk2Zh & 11.9 & 10.8 & 10.2 & 13.1 & 3.0 & 7.6 & \textbf{18.5} \\
Ko2Ja & 22.4 & 21.2 & 17.5 & 22.3 & 23.9 & 19.3 & \textbf{26.2} \\
Ko2Th & 11.9 & 9.7 & 11.2 & 12.2 & 0.8 & 2.0 & \textbf{14.7} \\
Ko2Zh & 20.0 & 20.3 & 10.2 & 16.2 & 16.6 & 14.7 & \textbf{22.7} \\
Th2Ar & 11.3 & 9.1 & 8.7 & 5.5 & 1.7 & 6.8 & \textbf{15.4} \\
Th2Es & 16.4 & 15.1 & 16.4 & 9.1 & 1.5 & 10.0 & \textbf{18.6} \\
Th2Fr & 22.2 & 20.3 & 19.3 & 12.4 & 5.1 & 14.2 & \textbf{25.2} \\
Th2Ja & 16.5 & 15.4 & 12.6 & 14.5 & 0.0 & 5.5 & \textbf{22.7} \\
Th2Kk & 2.2 & 1.7 & 4.4 & 5.4 & 0.3 & 0.8 & \textbf{9.3} \\
Th2Ko & 12.2 & 9.4 & 8.3 & 7.6 & 0.8 & 5.1 & \textbf{16.5} \\
Th2Zh & 18.8 & 18.0 & 9.5 & 9.2 & 0.0 & 6.4 & \textbf{22.3} \\
Tr2Ja & 17.7 & 7.3 & 11.9 & 20.0 & 19.6 & 21.4 & \textbf{23.2} \\
Uk2Fr & 27.5 & 24.3 & 31.2 & 33.0 & 31.5 & 33.0 & \textbf{34.6} \\
Uk2Ja & 17.3 & 13.0 & 14.1 & 20.1 & 16.9 & 21.8 & \textbf{26.6} \\
Uk2Kk & 3.4 & 2.7 & 7.9 & 8.3 & 0.7 & 1.8 & \textbf{12.4} \\
Uk2Ko & 13.2 & 8.9 & 10.7 & 15.8 & 16.0 & 17.3 & \textbf{18.9} \\
Uk2Th & 12.4 & 11.2 & 13.7 & 15.2 & 1.1 & 2.9 & \textbf{19.9} \\
Uk2Zh & 20.7 & 18.4 & 11.6 & 9.8 & 16.1 & 17.8 & \textbf{25.6} \\
Ur2Ar & 8.0 & 7.4 & 5.2 & 4.8 & 3.4 & 6.4 & \textbf{10.7} \\
Ur2Ko & 8.7 & 6.8 & 8.5 & 9.3 & 4.5 & 9.0 & \textbf{12.8} \\
Zh2Ar & 11.9 & 9.8 & 10.6 & 13.5 & 14.9 & 13.6 & \textbf{17.0} \\
Zh2Fr & 24.5 & 21.7 & 21.8 & 25.7 & 24.5 & 21.3 & \textbf{28.9} \\
Zh2Ja & 19.2 & 15.0 & 10.9 & 18.4 & 14.4 & 17.3 & \textbf{27.7} \\
Zh2Ko & 14.2 & 10.4 & 9.7 & 14.8 & 15.7 & 15.4 & \textbf{21.2} \\
Zh2Pt & 22.6 & 20.9 & 19.5 & 20.5 & 21.0 & 21.5 & \textbf{25.4} \\
Zh2Th & 13.3 & 10.8 & 12.8 & 13.9 & 1.0 & 2.5 & \textbf{19.8} \\
\midrule
Avg. Score & 14.6 & 11.9 & 12.2 & 13.9 & 9.7 & 11.9 & \textbf{19.7} \\
\bottomrule
\end{tabular}

}
\end{table}

\paragraph{English-pivot Translation Results}

The English-pivot translation results on Flores benchmark~\cite{flores101,nllb2022_flores200}, as detailed in Table~\ref{tab:mt_flores}, highlight the strengths of the Marco-Chat LLMs in translation tasks across a variety of languages. This benchmark is designed to evaluate translation quality from English to multiple target languages (EN$\to$XX) and vice versa (XX$\to$EN), offering insights into the multilingual capabilities of different models.

In the EN$\to$XX translation tasks, the Marco-72B model achieves an average score of 43.8, surpassing the second-best model, Google Translate, by a margin of 0.3 points. Notably, Marco-72B shows strong performance in translating English into Arabic (En2Ar) with a BLEU score of 61.2, outperforming Google by 11.2 points, and in Portuguese (En2Pt) with a BLEU score of 57.9, leading by 1.9 points. These results highlight the model's ability to handle both high-resource languages, such as French (En2Fr, 58.8), and low-resource languages, such as Ukrainian (En2Uk, 44.3), demonstrating its strong capability and robustness. Besides, for the language pairs where Marco-LLM underperformed competitive commercial MT systems such as En2De, En2Ja and En2Tr, it still achieved the best translation performance compared to the other open-sourced LLMs. For XX$\to$EN translations, Marco-72B achieves an impressive average score of 51.6, leading the performance with a significant margin of 8.0 points over the second-best model, Google. The model shows outstanding performance in translating from Italian (It2En, 49.2) and Korean (Ko2En, 49.0), reflecting its advanced capacity to capture nuanced linguistic features and deliver high-quality translations. The consistent outperformance across both high-resource languages (e.g., French to English, 56.8) and low-resource languages (e.g., Ukrainian to English, 58.8) underscores the effectiveness of our multilingual approach. The results on the Flores benchmark~\cite{nllb2022_flores200} validate the effectiveness of focusing on enhancing multilingual capabilities.

\paragraph{Non-English-pivot Translation Results}

The Non-English-pivot translation results~(\textit{Any2Any} translation - translation from any languages to any languages) the Flores benchmark, as shown in Table~\ref{tab:mt_any2any}, highlight the superior performance of the Marco-7B model in \textit{Any2Any} translation tasks.

The Marco-7B model achieves the highest average score of 19.5, which is a substantial margin above the second-best model, Qwen2-7B, with an average score of 14.4. This represents a notable improvement of 5.1 points. In some specific language directions, Marco-7B exhibits remarkable strong performance. For example, Marco-7B outperformed Qwen2-7B in Chinese to Japanese (Zh2Ja) translation by 8.5 points. Similarly, in the Arabic to Japanese (Ar2Ja) translation, Marco-7B achieves a score of 21.8, outperforming Llama3.1-8B by 4.3 points. Moreover, in the French to Japanese (Fr2Ja) translation, Marco-7B scores 25.8, which is 4.2 points higher than Llama3.1-8B. These results underscore the model's capability to manage complex linguistic structures, particularly in non-English-pivot language pairs. The Marco-7B model's superior performance is evident across both high-resource and low-resource languages, demonstrating its versatility and robustness. This is particularly important for enhancing the multilingual/cross-lingual capabilities of large language models, as it ensures consistent translation quality across a wide range of languages beyond traditional English-centric translation~\cite{fan2020englishcentricmultilingualmachinetranslation}.

\begin{figure}
    \centering
    \includegraphics[width=0.73\linewidth]{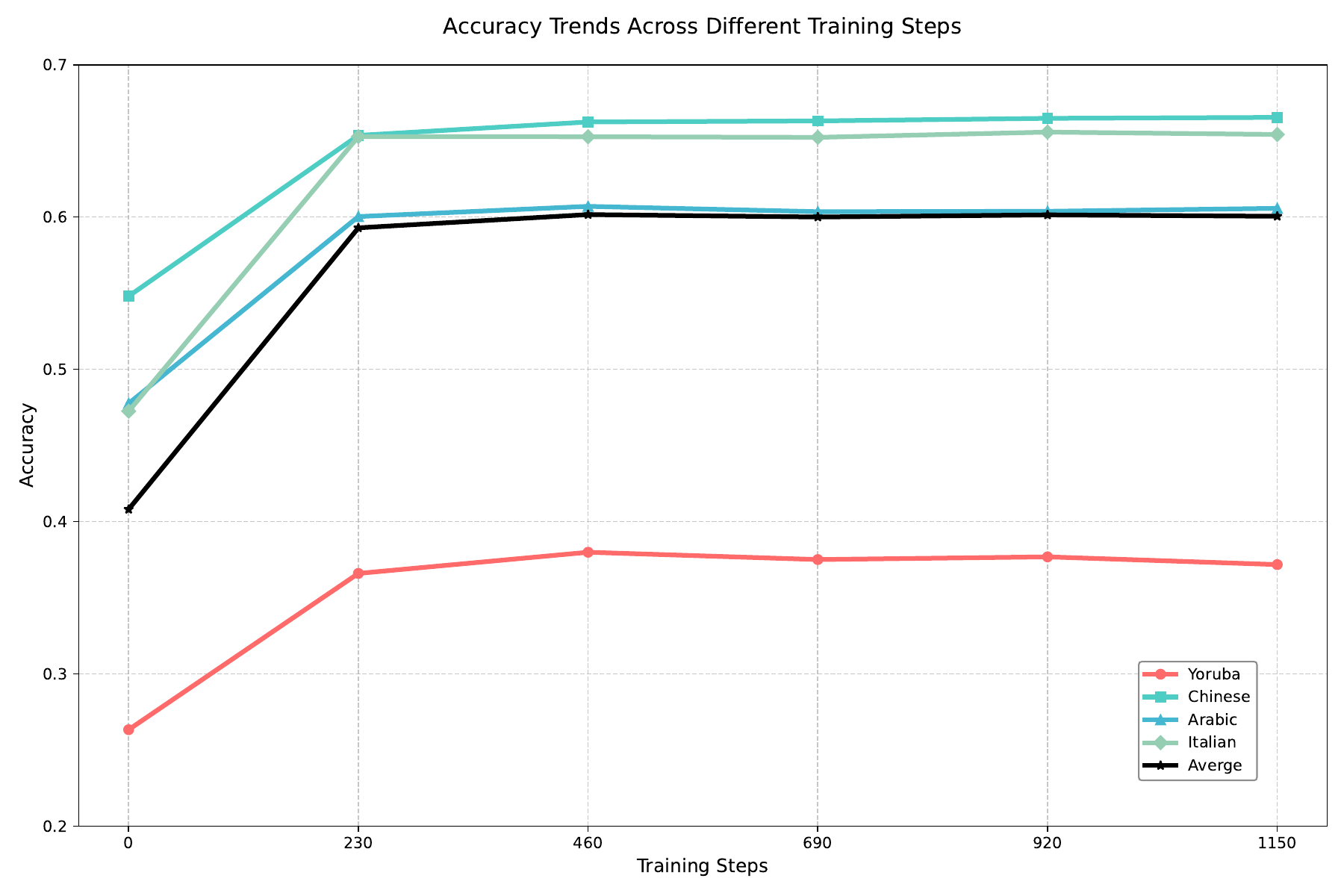}
\caption{Accuracy trends across training checkpoints for different languages on MMMLU. The model shows rapid initial learning (0-230 steps) followed by performance stabilization. High-resource languages (ZH-CN, IT-IT) consistently outperform low-resource ones (YO-NG), with a persistent performance gap of ~29\%.}    
\label{fig:mmmlu_checkpoint}
\end{figure}

\paragraph{Performance Across Different Training Steps}

Our analysis of the performance of Marco-7B model across different training steps (0-1150) on the MMMLU benchmark is shown in Figure~\ref{fig:mmmlu_checkpoint}. The most significant improvements occur during the early training phase (0-230 steps), where the overall average accuracy increases dramatically from 40.81\% to 59.29\%. After 460 step, the performance stabilizes across most languages, with the final overall accuracy reaching 60.05\%. High-resource languages like Chinese (ZH-CN) and Italian (IT-IT) achieve and maintain higher accuracy levels (>65\%) compared to low-resource languages, with Yoruba (YO-NG) plateauing around 37\%. This persistent performance gap of approximately 29\% between high and low-resource languages suggests that while the model effectively captures multilingual knowledge early in training, achieving better performance across languages remains a challenge that may require more monolingual data during the pretraining phase rather than simply extending training steps for SFT.

\subsection{Multilingual Preference Alignment}

Preference alignment is critical for ensuring that an LLM's outputs are consistent with human expectations and values. However, most LLMs are predominantly aligned on preference in English data, leading to a disparity in performance when applied to other languages. In a multilingual setting, this alignment becomes even more essential due to the variations in language structures~\cite{she-etal-2024-mapo}, idiomatic expressions, and cultural references. By focusing on multilingual preference alignment, we aim to enhance Marco's ability to generate responses that are not only grammatically correct but also culturally appropriate and contextually relevant in multiple languages.

Moreover, multilingual preference alignment helps mitigate biases that may arise from training on datasets that lack linguistic diversity. It promotes fairness and inclusivity, enabling the model to cater to a broader user base. By aligning the model's preferences across different languages, we ensure that Marco-LLM can effectively understand and respond to users worldwide, fostering better communication and understanding across linguistic boundaries.

\subsubsection{Dataset Construction from Existing Preference Data}
To construct a comprehensive multilingual preference dataset, we began with the \textit{LMSYS Arena Human Preference} dataset~\cite{chiang2024chatbot}~\footnote{\url{https://huggingface.co/datasets/lmsys/lmsys-arena-human-preference-55k}}. This dataset comprises 57.5k high-quality human preference annotations for various prompts and responses in English. We selected a subset of high-quality examples based on criteria such as clarity, relevance, and diversity of topics to ensure a robust foundation for multilingual preference alignment. The selected examples were then translated into the 28 target languages. This translation step was critical to extend the language coverage of the original data. By leveraging existing English preference data and extending it to multiple languages, we aim to improve the performance of preference alignment of Marco-LLM under various languages beyond English.

\subsubsection{Multilingual Preference Data Generation and Translation}
In addition to the translated data, we expanded our preference dataset by incorporating prompts from the \textit{UltraFeedback} dataset \cite{cui2023ultrafeedback}, which are also translated into 28 languages. For each prompt, we utilized Marco-LLM to generate at least two distinct responses with different generation configuration. This approach allowed us to capture the model's inherent variability in generating responses across different languages. To establish preferences between the generated responses, we employed another LLM to evaluate and select the better response based on predefined criteria such as relevance, coherence, and adherence to the prompt. This process effectively created a set of preference pairs that reflect the model's capabilities and the desired outcomes in various languages. By generating and evaluating responses within the target languages, we ensured that the preference data was culturally and linguistically appropriate.

\subsubsection{Evaluation Results}
To evaluate Marco's multilingual capabilities for capturing preference in different languages, we translated the original English MT-Bench benchmark~\cite{chiang2024chatbot} into the 28 target languages. We then compare the generated responses from LLMs in a pairwise manner using GPT-4o-mini, specifically we compare the responses of Marco-LLM (7B) with the responses from the other six baseline LLMs including Qwen2, Qwen2.5, Llama3, Llama3.1, Aya-23, Aya-expanse (all in 7/8B size).

\begin{figure}
    \centering
    \includegraphics[width=\linewidth]{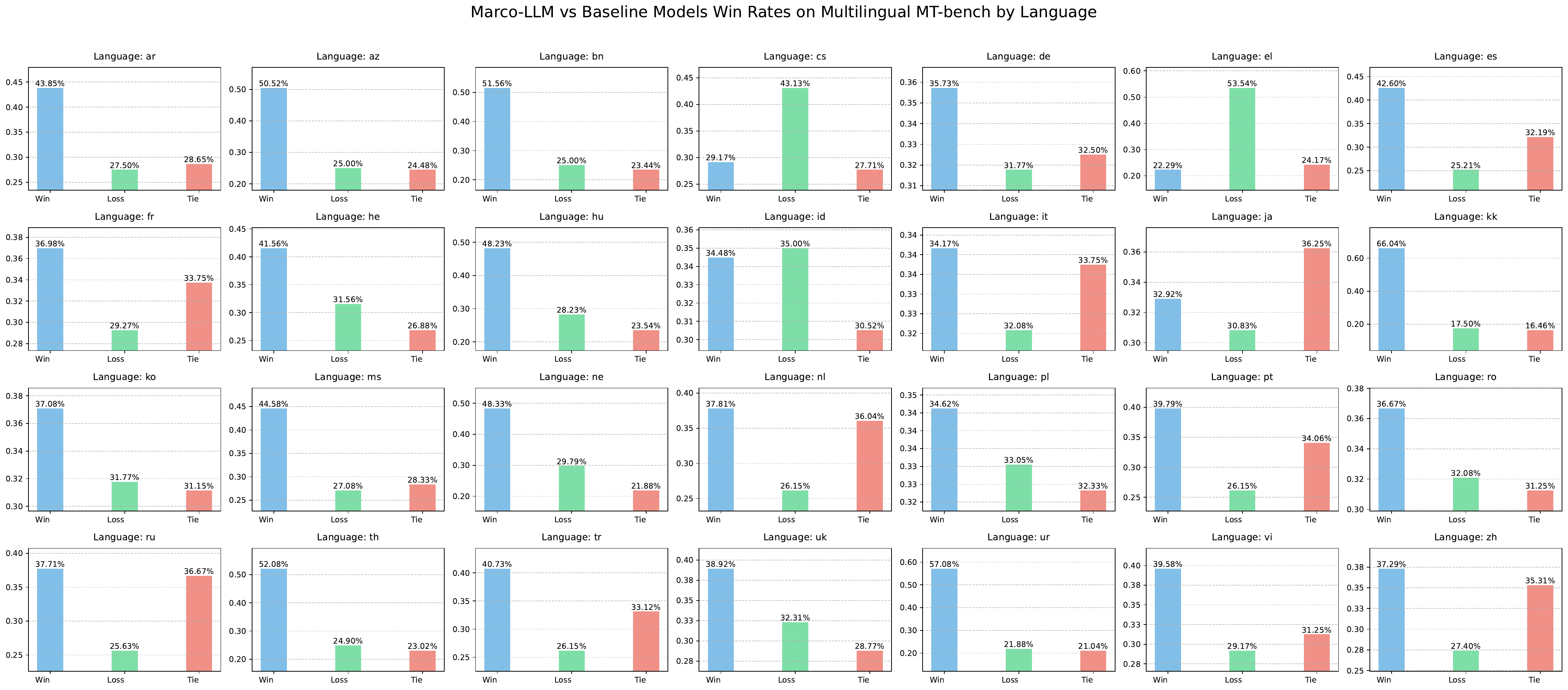}
    \caption{Performance comparison of Marco-LLM against baseline models across 28 languages on multilingual MT-bench. Each subplot shows the win rate (\textcolor{blue}{blue}), loss rate (\textcolor{green}{green}), and tie rate (\textcolor{red}{red}) for a specific language. Win rates indicate Marco-LLM's superior responses, loss rates represent baseline models' better performance, and tie rates show equivalent quality responses.}
    \label{fig:mt_bench}
\end{figure}

The results from the multilingual MT-bench, as illustrated in Figure~\ref{fig:mt_bench}, reveal that Marco-chat-7B model outperforms baseline models in 25 out of 28 languages. We evaluate model responses using GPT-4o-mini where win rates, loss rates, and tie rates (Marco-LLM vs baseline) were averaged for the baseline models mentioned in Section~\ref{sec:baseline_models_post_training} across each language. Marco-chat-7B achieved better generation quality in low-resource languages. For instance, in Azerbaijani (\textit{az}), the model achieves a win rate of 50.52\% compared to a loss rate of 25\%, while in Bengali (\textit{bn}), the win rate is 51.56\% against a loss rate of 23.44\% as well as Kazakh(\textit{he}) where Marco-LLM obtained a win rate of 66.04\% . These results indicate a clear advantage over the baseline models. Hebrew (\textit{he}) also demonstrates a strong performance with a win rate of 41.56\%, surpassing the loss rate of 31.56\%. In certain high-resource languages, Marco-chat-7B maintains competitive performance. For example, in French (\textit{fr}), our model achieves a win rate of 35.98\% against a loss rate of 29.27\%, and in Chinese (\textit{zh}), it achieves a win rate of 37.29\% compared to a loss rate of 27.40\%. These results reflect the model’s effective handling of languages with extensive linguistic data. The model also performs consistently well across various language families, maintaining win rates around 35\% for Indo-European languages such as Italian (\textit{it}) with a win rate of 34.17\%, and Dutch (\textit{nl}) with a win rate of 37.81\%. These results suggest balanced performance across diverse linguistic structures. While Marco-LLM achieved strong performance in 25 languages out of 28 languages, it still underperformed in languages including Czech, Greek and Indonesian. This suggests the need for further refinement in handling complex grammatical structures. Overall, the experimental results highlight Marco-chat-7B's strong multilingual performance, particularly in languages where it achieves a higher win rate than loss rate. The model effectively addresses the challenges posed by both high-resource and low-resource languages, demonstrating its capability and potential for deployment in a wide range of linguistic environments.

\section{Conclusion and Future Work}

\label{sec:conclusion}

In this paper, we introduced Marco-LLM, a multilingual LLM specifically designed to address the challenges posed by low-resource languages. By leveraging a large and diverse multilingual dataset, we conducted extensive multilingual continual pre-training and post-training, including supervised finetuning and preference alignment, based on the Qwen2 model. Our comprehensive evaluations on benchmarks such as MMMLU, Flores, Belebele, CEVAL, TydiQA, and multilingual MT-bench validated that Marco-LLM obtained excellent performance in multilingual tasks. The results demonstrate that focusing on low-resource languages can bridge existing performance gaps and extend the benefits of LLMs to a wider range of linguistic communities. Our work highlights the importance of data diversity and targeted training strategies in enhancing model performance across diverse languages. For future work, there are several directions for future research. One promising direction is to extend Marco-LLM's capabilities to include more languages, further enriching the linguistic diversity it can handle. Additionally, exploring the integration of multilingual reasoning capabilities could enhance the model's ability to understand and generate more complex language structures. Furthermore, improving model efficiency and scalability will be essential for deploying these systems in real-world applications, particularly in resource-constrained environments.

\bibliography{references}

\appendix 
\section{Appendix}
\label{sec:appendix_data}

\subsection{Dataset Preprocessing}
\subsubsection{Translation Templates}\label{app:translation_templates}

\begin{table}[htbp]
    \centering
    \caption{Translation templates used in our experiments.        \textit{Note:} The translation templates are used to construct our parallel data, where $\Diamond$ indicates the position of a line break. The placeholders \texttt{<src\_lang>}, \texttt{<tgt\_lang>}, \texttt{<input>}, and \texttt{<output>} represent the source language name, target language name, source text, and target text in the parallel pair, respectively.}
    \label{tab:translation_templates}
    \resizebox{\linewidth}{!}{%
        \begin{tabular}{@{}ll@{}}
            \toprule
            \textbf{ID} & \textbf{Template (in English)} \\
            \midrule
            A & \texttt{<src\_lang> phrase: <input> $\Diamond$ <tgt\_lang> phrase: <output>} \\
            B & \texttt{<src\_lang> text: <input> $\Diamond$ <tgt\_lang> text: <output>} \\
            C & \texttt{Translate the text from <src\_lang> to <tgt\_lang>: $\Diamond$ <src\_lang> text: <input> $\Diamond$ <tgt\_lang> text: <output>} \\
            D & \texttt{Translate the words from <src\_lang> to <tgt\_lang>: $\Diamond$ <src\_lang> words: <input> $\Diamond$ <tgt\_lang> words: <output>} \\
            E & \texttt{Convert the phrase from <src\_lang> to <tgt\_lang>: $\Diamond$ <src\_lang> phrase: <input> $\Diamond$ <tgt\_lang> phrase: <output>} \\
            F & \texttt{Render the <src\_lang> sentence <input> to <tgt\_lang>: <output>} \\
            G & \texttt{Provide the translation of the sentence <input> from <src\_lang> to <tgt\_lang>: <output>} \\
            H & \texttt{Change the phrase <input> to <tgt\_lang>, the translated phrase is: <output>} \\
            I & \texttt{Please change the sentence <input> to <tgt\_lang>, and the resulting translation is: <output>} \\
            J & \texttt{Change the phrase <input> to <tgt\_lang>, resulting in: <output>} \\
            K & \texttt{The sentence <input> in <src\_lang> means <output> in <tgt\_lang>} \\
            \bottomrule
        \end{tabular}
    }

\end{table}

\begin{table}[htbp]
    \centering
    \caption{Prompt Templates for Synthetic Data.}
    \label{tab:synthetic_prompts}
    \scalebox{0.8}{
    \begin{tabular}{@{}p{5.5cm}p{15cm}@{}}
        \toprule[1pt]
        \textbf{Method} & \textbf{Prompt} \\
        \midrule[1pt]
        \textbf{Keywords-based Explanation} & 
\setlength{\fboxrule}{0pt}

        \fbox{\parbox{\linewidth}{%
            \textit{Suppose that you are a/an} \texttt{\{role\_1\}} \textit{in} \texttt{\{subject\}}. \textit{Please explain the following keywords and meet the following requirements:}
            \begin{enumerate}[leftmargin=1.5em, label=(\arabic*)]
                \item \textit{The keywords:} \texttt{\{keywords\}}\textit{;}
                \item \textit{Each keyword explanation should contain at least three sentences. You can generate a story about the keyword for better explanation;}
                \item \textit{The explanations suit} \texttt{\{role\_2\}} \textit{students;}
                \item \textit{Summarize the explanations.}
            \end{enumerate}
            \textit{Your answer should be a list of keywords. Make the explanations correct, useful, understandable, and diverse.}
        }}
        \\[1ex]
        \midrule[1pt]
        \textbf{Keywords-based Story} & 
\setlength{\fboxrule}{0pt}
        \fbox{\parbox{\linewidth}{%
            \textit{Assume that you are a/an} \texttt{\{role\_1\}} \textit{in} \texttt{\{subject\}}. \textit{Before you teach students new vocabulary, please write a} \texttt{\{type\_passage\}} \textit{about the new knowledge and meet the following requirements:}
            \begin{enumerate}[leftmargin=1.5em, label=(\arabic*)]
                \item \textit{It must contain keywords:} \texttt{\{keywords\}}\textit{;}
                \item \textit{Its setting should be} \texttt{\{scene\}}\textit{;}
                \item \textit{Should be between} \texttt{\{min\_length\}} \textit{and} \texttt{\{max\_length\}} \textit{words in length;}
                \item \textit{The writing style should be} \texttt{\{style\}}\textit{;}
                \item \textit{The suitable audience is} \texttt{\{role\_2\}}\textit{;}
                \item \textit{Should end with} \texttt{\{ending\}}\textit{;}
                \item \textit{Should be written in} \texttt{\{language\}}\textit{.}
            \end{enumerate}
        }}
        \\[1ex]
        \midrule[1pt]
        \textbf{Few-shot Based SFT Data} & 
\setlength{\fboxrule}{0pt}

        \fbox{\parbox{\linewidth}{%
            \textit{I want you to act as a Sample Generator. Your goal is to draw inspiration from the} \texttt{Given Sample} \textit{to create a brand new sample. This new sample should belong to the same domain as the} \texttt{Given Sample} \textit{but be even rarer. The length and complexity of the} \texttt{Created Sample} \textit{should be similar to that of the} \texttt{Given Sample}\textit{. The} \texttt{Created Sample} \textit{must be reasonable and understandable by humans. The terms} \texttt{Given Sample}\textit{,} \texttt{Created Sample}\textit{, 'given sample', and 'created sample' are not allowed to appear in the} \texttt{Created Sample}\textit{.}

            \vspace{1em}
            \texttt{Given Sample:}
            \begin{enumerate}[leftmargin=1.5em, label=(\arabic*)]
                \item \textit{Sample doc 1}
                \item \textit{Sample doc 2}
                \item \textit{Sample doc 3}
                \item \textit{...}
            \end{enumerate}

            \vspace{0.5em}
            \texttt{Created Sample:}
        }}
        \\
        \bottomrule[1pt]
    \end{tabular}
    }
\end{table}

To standardize the translation process, we designed a diverse set of translation templates as shown in Table~\ref{tab:translation_templates}. These templates serve multiple purposes: they provide consistent formatting for the parallel data, enable clear instruction-following capabilities, and help maintain structural consistency across different language pairs. The templates range from simple direct translation formats (Templates A and B) to more elaborate instructional patterns (Templates C through K), each designed to capture different aspects of the translation task. By incorporating various phrasings and structures, these templates help improve the model's robustness and ability to handle diverse translation requests. The symbol $\Diamond$ in the templates is a line break, which helps maintain clear visual separation between different parts of the translation pair.

\subsubsection{Prompt Templates}\label{app:prompt_templates}
Table \ref{tab:synthetic_prompts} summaries the prompt templates for synthetic data.

\subsection{More Evaluation Results for Instruct LLMs}

We show more evaluation results for the post-trained LLMs in this section. In our evaluation on the TydiQA benchmark~\cite{clark-etal-2020-tydiqa} shown in Table~\ref{tab:tydiqa_results}, Marco-Chat consistently exhibits superior performance across both 7B and 70B model categories, underlining its enhanced multilingual capabilities. Notably, Marco-Chat achieves top scores in languages such as Arabic and Bengali, demonstrating its proficiency in managing diverse linguistic structures and dialects. With an impressive average score of 57.7 for 7B models and 61.0 for 70B models, Marco-Chat surpasses competitors, showcasing its robust understanding across many languages. The CEVAL benchmark~\cite{huang2023ceval} results in Table~\ref{tab:sft_ceval_results} underscore Marco's strong performance across both 7B and 70B models. Marco-LLM consistently achieves the highest scores, with an average of 86.4 in the 7B models and 94.5 in the 70B models, demonstrating its robust generalization across diverse linguistic tasks. Particularly noteworthy is Marco's strong performance in complex categories such as 'Hard' and 'STEM', where it obtained higher scores at handling challenging language tasks and quantitative reasoning. The AGIEval benchmark~\cite{zhong2023agievalhumancentricbenchmarkevaluating} results shown in Table~\ref{tab:agieval_results}highlight Marco-Chat's exceptional multilingual and reasoning capabilities, particularly in the 7B model category, where it obtained strong performance in Chinese and Gaokao-related tasks. Achieving top scores in categories such as Gaokao-English and Gaokao-History underscores Marco-Chat's adeptness at handling diverse linguistic challenges and contextual comprehension. In the 70B model category, Marco-Chat continues to lead with an average score of 72.7.

\begin{table}[!htb]
\centering
\caption{Performance comparison of language models on TydiQA benchmark~\cite{clark-etal-2020-tydiqa} across different languages.}
\label{tab:tydiqa_results}
\scalebox{0.8}{
\begin{tabular}{l rrrr rrr}
\toprule

\textbf{Language} & \textbf{Qwen2} & \textbf{Qwen2.5} & \textbf{Llama-3} & \textbf{Llama3.1} & \textbf{Aya-23} & \textbf{Aya-expanse} & \textbf{Marco-Chat} \\
\midrule
        \rowcolor{lightgray} \multicolumn{8}{l}{\textbf{7B Models}} \\

        Arabic & 47.5 & 48.0 & 58.8 & 68.3 & 67.2 & 45.0 & \textbf{78.4} \\
Bengali & 42.3 & 61.6 & 56.8 & 64.3 & 50.6 & 21.3 & \textbf{74.6} \\
English & 29.0 & 42.7 & 24.4 & 41.6 & 53.6 & 49.6 & \textbf{44.4} \\
Finnish & 27.1 & 48.4 & 54.8 & 47.3 & 44.7 & 23.5 & \textbf{51.5} \\
Indonesian & 28.6 & 38.4 & 34.5 & 31.8 & 33.4 & 30.2 & \textbf{48.1} \\
Japanese & 0.8 & 8.7 & 17.6 & 31.8 & \textbf{71.0} & 41.7 & 70.5 \\
Korean & 38.2 & 45.6 & 25.4 & 57.9 & {76.0} & 20.3 & \textbf{77.9} \\
Russian & 22.4 & 33.7 & 31.2 & 40.8 & \textbf{49.9} & 25.7 & {46.6} \\
Swahili & 17.5 & 8.9 & 30.1 & 45.2 & 11.8 & 13.3 & \textbf{31.3} \\
Telugu & 28.7 & 22.8 & 48.8 & \textbf{83.3} & 5.5 & 11.3 & 35.2 \\
Thai & 39.7 & 55.7 & 54.3 & 70.8 & 55.2 & 29.3 & \textbf{76.2} \\
\midrule
Avg. Scores & 29.2 & 39.0 & 39.7 & 53.0 & 47.2 & 28.3 & \textbf{57.7} \\
\midrule
        \rowcolor{lightgray} \multicolumn{8}{l}{\textbf{70B Models}} \\
Arabic & 42.5 & 61.6 & 72.7 & 62.4 & 67.5 & 40.9 & \textbf{78.4} \\
Bengali & 47.1 & 65.7 & 55.9 & 68.1 & 43.6 & 28.0 & \textbf{73.3} \\
English & 43.0 & 45.5 & 46.6 & 41.2 & 53.6 & 46.8 & \textbf{51.2} \\
Finnish & 55.8 & 39.0 & 57.0 & 50.9 & 52.5 & 26.0 & \textbf{57.0} \\
Indonesian & 39.9 & \textbf{47.8} & 44.0 & 31.9 & 35.8 & 27.6 & 46.5 \\
Japanese & 38.6 & 44.7 & 43.8 & 49.9 & \textbf{73.6} & 22.1 & 68.4 \\
Korean & 41.1 & 43.2 & 43.8 & 63.0 & \textbf{73.1} & 29.5 & 69.4 \\
Russian & 38.5 & 43.2 & 42.5 & 36.3 & 52.4 & 35.5 & \textbf{45.9} \\
Swahili & 18.6 & 21.3 & 46.0 & 33.8 & 20.4 & 12.5 & \textbf{49.2} \\
Telugu & 36.9 & 41.4 & 45.5 & \textbf{80.4} & 26.6 & 22.9 & 62.8 \\
Thai & 41.5 & 68.2 & 74.0 & 66.4 & 52.7 & 38.7 & \textbf{76.5} \\
\midrule
Avg. Scores & 40.3 & 48.4 & 52.0 & 53.1 & 50.2 & 30.0 & \textbf{61.0} \\

\bottomrule
\end{tabular}
}
\end{table}

\begin{table}[!htb]
\centering
\caption{Performance comparison on the CEVAL benchmark across different categories.}
\label{tab:sft_ceval_results}
\scalebox{0.8}{
\begin{tabular}{l rrr rrrr}
\toprule
\textbf{Category} & \textbf{Qwen2} & \textbf{Qwen2.5} & \textbf{Llama3} & \textbf{Llama3.1} & \textbf{Aya-23} & \textbf{Aya-expanse} & \textbf{Marco} \\
\midrule
\rowcolor{lightgray} \multicolumn{8}{l}{\textbf{7B Models}} \\
Average & 81.8 & 77.9 & 50.8 & 55.6 & 43.9 & 48.5 & \textbf{86.4} \\
Hard & 63.1 & 51.8 & 33.9 & 36.9 & 32.6 & 32.4 & \textbf{79.9} \\
Other & \textbf{84.9} & 82.8 & 53.5 & 56.3 & 44.9 & 48.3 & 81.0 \\
Humanities & 84.3 & 79.6 & 47.2 & 53.8 & 43.2 & 50.3 & \textbf{84.8} \\
Social Science & \textbf{91.3} & 87.8 & 58.9 & 64.9 & 51.2 & 54.3 & 90.4 \\
STEM & 73.9 & 69.4 & 47.2 & 51.5 & 40.1 & 44.7 & \textbf{88.3} \\
\midrule
\rowcolor{lightgray} \multicolumn{8}{l}{\textbf{70B Models}} \\
Average & 90.6 & 88.2 & 66.7 & 71.6 & 53.6 & 56.9 & \textbf{94.5} \\
Hard & 77.8 & 73.3 & 51.1 & 56.0 & 35.2 & 36.5 & \textbf{94.0} \\
Other & 92.0 & 88.4 & 63.1 & 69.2 & 52.8 & 54.7 & \textbf{92.5} \\
Humanities & 92.8 & 91.0 & 66.5 & 70.0 & 56.9 & 60.3 & \textbf{95.2} \\
Social Science & 94.8 & 91.5 & 75.6 & 82.0 & 61.8 & 66.4 & \textbf{95.7} \\
STEM & 88.4 & 84.9 & 64.2 & 68.5 & 48.1 & 51.4 & \textbf{94.6} \\
\bottomrule
\end{tabular}
}
\end{table}

\begin{table}[!htb]
\centering
\caption{Agieval 7B Results: Performance of various models across different categories of the Agieval dataset. The best performance in each category is highlighted in bold.}
\label{tab:agieval_results}
\scalebox{0.8}{
\begin{tabular}{l rrrrr rrrrr rrrrr rrrrr rrr}
\toprule
\textbf{Category} & \textbf{Qwen2} & \textbf{Qwen2.5} & \textbf{Llama3} & \textbf{Llama3.1} & \textbf{Aya-23} & \textbf{Aya-expanse} & \textbf{Marco-Chat} \\
\midrule
        \rowcolor{lightgray} \multicolumn{8}{l}{\textbf{7B Models}} \\

Chinese & 59.5 & 57.1 & 37.7 & 62.5 & 36.5 & 34.4 & \textbf{65.3} \\
English & 54.0 & \textbf{61.5} & 41.5 & 51.0 & 37.9 & 39.9 & 56.6 \\
Gaokao & 64.5 & 63.4 & 41.5 & 53.0 & 40.3 & 37.3 & \textbf{71.0} \\
Gaokao-Chinese & 77.6 & 67.9 & 47.6 & 58.5 & 46.3 & 42.3 & \textbf{80.5} \\
Gaokao-English & 89.5 & 85.3 & 79.1 & 89.5 & 82.0 & 67.0 & \textbf{93.1} \\
Gaokao-Geography & 81.4 & 80.4 & 49.8 & 75.9 & 50.8 & 50.3 & \textbf{86.9} \\
Gaokao-History & 86.0 & 82.1 & 55.7 & 74.9 & 51.1 & 49.8 & \textbf{89.8} \\
Gaokao-Biology & 81.9 & 80.5 & 52.9 & 76.2 & 43.8 & 35.7 & \textbf{87.6} \\
Gaokao-Chemistry & 56.5 & 55.6 & 32.4 & 46.9 & 29.5 & 30.4 & \textbf{74.4} \\
Gaokao-MathQA & 45.3 & 55.3 & 32.5 & 16.2 & 27.4 & 27.4 & \textbf{60.7} \\
Gaokao-Physics & 45.0 & 44.5 & 15.5 & 34.0 & 30.5 & 26.0 & \textbf{57.5} \\
Gaokao-MathCloze & 17.0 & \textbf{18.6} & 8.5 & 5.1 & 1.7 & 6.8 & 8.5 \\
LogiQA-ZH & 60.8 & 54.7 & 42.1 & 61.1 & 36.3 & 37.8 & \textbf{68.2} \\
LSAT-AR & 26.5 & 23.9 & 23.5 & \textbf{30.9} & 21.3 & 23.0 & 29.6 \\
LSAT-LR & 57.3 & 66.3 & 54.5 & \textbf{84.5} & 41.0 & 45.3 & 79.2 \\
LSAT-RC & 66.2 & 74.4 & 69.5 & \textbf{89.2} & 55.8 & 56.1 & 75.8 \\
LogiQA-EN & 46.7 & 50.4 & 43.5 & 59.6 & 36.7 & 35.3 & \textbf{61.4} \\
SAT-Math & 82.7 & \textbf{88.6} & 68.2 & 44.1 & 35.9 & 49.1 & 62.3 \\
SAT-EN & 83.0 & \textbf{84.5} & 82.0 & 91.3 & 77.2 & 64.1 & 84.0 \\
SAT-EN-Without-Passage & 46.6 & 42.2 & 46.1 & 44.7 & 42.2 & 39.8 & \textbf{52.9} \\
Math & 82.7 & 48.5 & 20.0 & \textbf{57.6} & 7.1 & 13.6 & 12.8 \\
Aqua-RAT & 59.8 & \textbf{74.8} & 52.0 & 61.0 & 24.0 & 32.7 & 51.2 \\
JEC-QA-KD & \textbf{38.0} & 33.1 & 17.5 & 26.9 & 18.2 & 17.8 & 37.5 \\
JEC-QA-CA & 34.7 & 27.4 & 19.1 & 26.0 & 20.4 & 21.1 & \textbf{38.4} \\
\midrule
Average & 57.1 & 59.0 & 43.4 & 41.8 & 37.1 & 36.7 & \textbf{61.5} \\
\midrule
        \rowcolor{lightgray} \multicolumn{8}{l}{\textbf{70B Models}} \\
\midrule
Chinese & 66.3 & 66.0 & 51.0 & 49.3 & 43.5 & 42.1 & \textbf{74.5} \\
English & 65.7 & 69.4 & 65.3 & 62.5 & 45.5 & 50.6 & \textbf{70.4} \\
Gaokao & 71.8 & 71.9 & 54.1 & 53.0 & 47.6 & 46.6 & \textbf{80.6} \\
Gaokao-Chinese & 85.0 & 84.6 & 56.5 & 58.5 & 50.8 & 52.4 & \textbf{92.7} \\
Gaokao-English & 89.5 & 92.2 & 90.9 & 89.5 & 88.9 & 71.6 & \textbf{96.1} \\
Gaokao-Geography & 88.9 & 87.9 & 74.9 & 75.9 & 66.3 & 65.8 & \textbf{96.0} \\
Gaokao-History & 92.3 & 87.2 & 71.1 & 74.9 & 68.9 & 63.4 & \textbf{95.3} \\
Gaokao-Biology & 86.2 & 86.2 & 74.3 & 76.2 & 54.8 & 53.3 & \textbf{95.2} \\
Gaokao-Chemistry & 68.1 & 66.2 & 39.6 & 46.9 & 37.2 & 35.8 & \textbf{84.1} \\
Gaokao-MathQA & 60.7 & \textbf{66.1} & 45.0 & 16.2 & 29.1 & 36.8 & 74.6 \\
Gaokao-Physics & 53.5 & 56.5 & 23.0 & 34.0 & 31.5 & 32.0 & \textbf{75.0} \\
Gaokao-MathCloze & \textbf{22.0} & 20.3 & 11.9 & 5.1 & 0.9 & 8.5 & 16.1 \\
LogiQA-ZH & 70.2 & 70.1 & 61.1 & 61.1 & 46.2 & 46.2 & \textbf{82.3} \\
LSAT-AR & 32.2 & 27.4 & 31.7 & 30.9 & 19.6 & 20.9 & \textbf{40.0} \\
LSAT-LR & 73.1 & 88.0 & 80.8 & 84.5 & 60.0 & 59.4 & \textbf{94.5} \\
LSAT-RC & 77.7 & 85.5 & 84.8 & 89.2 & 72.9 & 62.1 & \textbf{91.1} \\
LogiQA-EN & 56.8 & 59.6 & 55.8 & 59.6 & 43.9 & 40.3 & \textbf{75.7} \\
SAT-Math & \textbf{90.5} & 84.1 & 86.4 & 44.1 & 43.6 & 66.8 & 74.6 \\
SAT-EN & 89.8 & \textbf{91.3} & 89.8 & \textbf{91.3} & 87.9 & 77.7 & 90.8 \\
SAT-EN w/o Passage & 49.0 & 49.5 & 55.3 & 44.7 & 44.7 & 45.6 & \textbf{69.4} \\
Math & 46.5 & \textbf{63.1} & 34.2 & 57.6 & 8.9 & 25.7 & 22.6 \\
AQUA-RAT & 75.2 & \textbf{76.0} & 69.3 & 61.0 & 28.0 & 32.0 & 75.2 \\
JEC-QA-KD & 39.0 & 35.5 & 34.4 & 26.9 & 23.2 & 19.3 & \textbf{41.4} \\
JEC-QA-CA & 39.9 & 20.3 & 28.9 & 26.0 & 24.1 & 19.5 & \textbf{44.6} \\
\midrule
Avg. Scores & 66.0 & 67.5 & 57.1 & 55.0 & 44.4 & 45.7 & \textbf{72.7} \\
\bottomrule
\end{tabular}
}
\end{table}

\end{document}